\documentclass{article}

\usepackage{microtype}
\usepackage{graphicx}
\usepackage{subcaption}
\usepackage{booktabs}

\usepackage{hyperref}

\usepackage[preprint]{icml2026}

\usepackage{amsmath}
\usepackage{amssymb}
\usepackage{mathtools}
\usepackage{amsthm}
\usepackage{overpic}
\usepackage{multirow}
\usepackage[table]{xcolor} 
\usepackage{enumitem}
\usepackage{makecell}
\usepackage{bm}
\usepackage{wrapfig}
\usepackage{amsmath}
\usepackage{amssymb}
\usepackage{caption}
\usepackage{overpic}
\usepackage{pifont}
\definecolor{commentgray}{RGB}{85, 85, 85}

\usepackage[capitalize,noabbrev]{cleveref}

\theoremstyle{plain}

\theoremstyle{definition}

\theoremstyle{remark}

\def\ie{\textit{i.e.}}
\def\eg{\textit{e.g.}}

\newcommand{\figref}[1]{Fig.~\ref{#1}}

\newcommand{\secref}[1]{Sec.~\ref{#1}}
\newcommand{\tableref}[1]{Tab.~\ref{#1}}

\newcommand{\dataset}{\textit{\textbf{PEBench}}}
\newcommand{\name}{{\textit{\textbf{World-Shaper}}}}

\usepackage[textsize=tiny]{todonotes}

\icmltitlerunning{World-Shaper: A Unified Framework for 360\textdegree{} Panoramic Editing}

\begin{document}

\twocolumn[
  \icmltitle{\name: A Unified Framework for 360\textdegree{} Panoramic Editing}
  \icmlsetsymbol{equal}{*}

\begin{icmlauthorlist}
\icmlauthor{Dong Liang*}{tj,cityu}  
\icmlauthor{Yuhao Liu*}{cityu}      
\icmlauthor{Jinyuan Jia}{hkust,tj}
\icmlauthor{Youjun Zhao}{cityu}
\icmlauthor{Rynson W. H. Lau}{cityu}
\end{icmlauthorlist}

\icmlaffiliation{tj}{Tongji University}
\icmlaffiliation{cityu}{CityUHK}
\icmlaffiliation{hkust}{HKUST(GZ)}

\icmlcorrespondingauthor{Jinyuan Jia}{jinyuanjia@hkust-gz.edu.cn}
\icmlcorrespondingauthor{Rynson W. H. Lau}{Rynson.Lau@cityu.edu.hk}

\icmlkeywords{Machine Learning, ICML}

\vskip 0.2in

\begin{center}
\centerline{\includegraphics[width=1\textwidth]{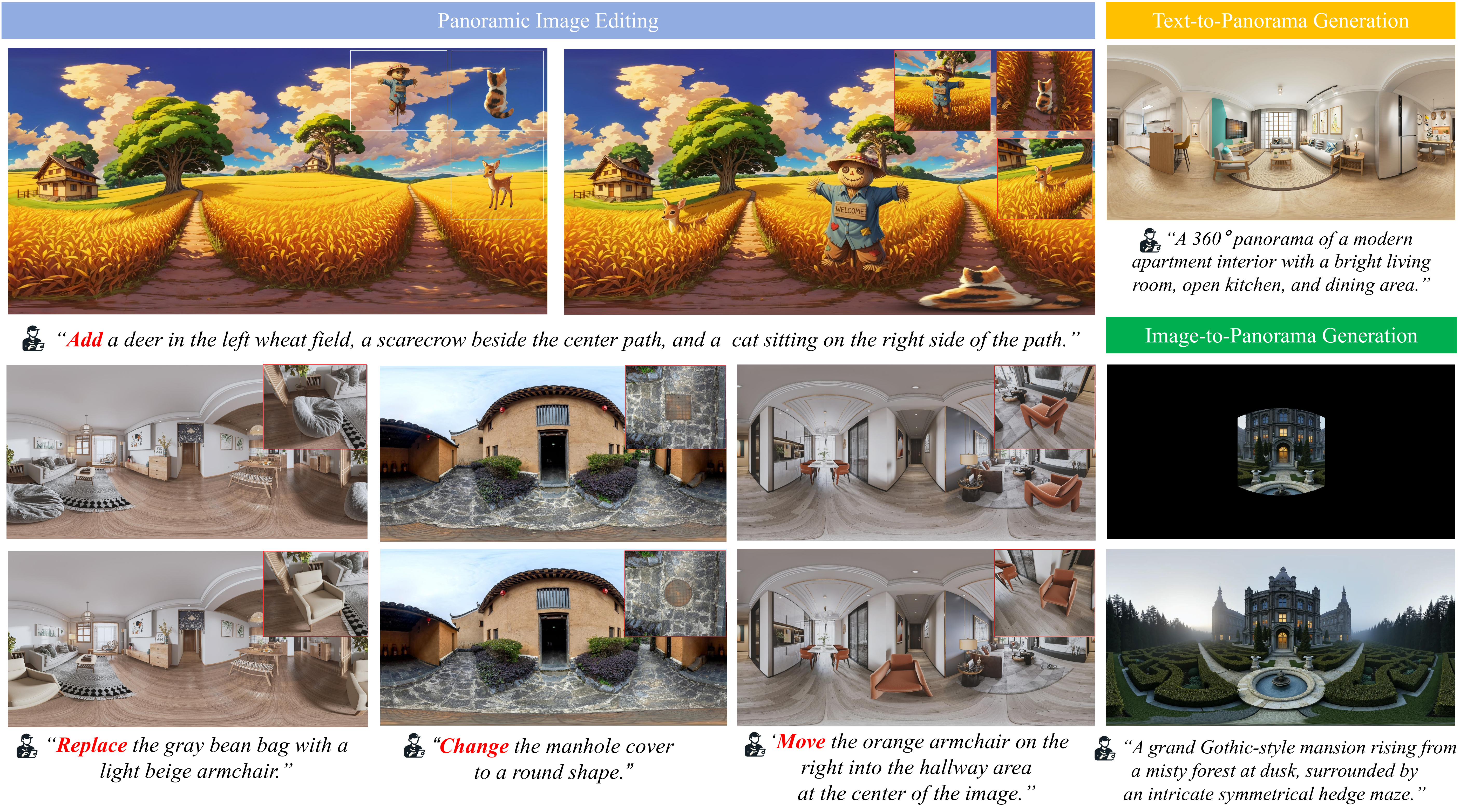}}
\captionof{figure}{Our geometry-aware diffusion framework, \name, unifies panorama generation and editing within a single editing-centric model, supporting  text-to-panorama and image-to-panorama generation, and diverse editing operations such as object addition, removal, replacement, relocation, and appearance modification. New objects and edited regions are highlighted in white and red boxes, respectively.}
\label{fig:teaser}
\end{center}

]

\printAffiliationsAndNotice{\icmlEqualContribution}

\begin{abstract}

Being able to edit panoramic images is crucial for creating realistic 360\textdegree{} visual experiences. However, existing perspective-based image editing methods fail to model the spatial structure of panoramas. 
Conventional cube-map decompositions attempt to overcome this problem but inevitably break global consistency due to their mismatch with spherical geometry. Motivated by this insight, we reformulate panoramic editing directly in the equirectangular projection (ERP) domain and present \name, a unified geometry-aware  framework that bridges panoramic generation and editing within a single editing-centric design. To overcome the scarcity of paired data, we adopt a generate-then-edit paradigm, where controllable panoramic generation serves as an auxiliary stage to synthesize diverse paired examples for supervised editing learning. To address geometric distortion, we introduce a geometry-aware learning strategy that explicitly enforces position-aware shape supervision and implicitly internalizes panoramic priors through progressive training. Extensive experiments on our new benchmark, \dataset, demonstrate that \name ~achieves superior geometric consistency, editing fidelity, and text controllability compared to SOTA methods, enabling coherent and flexible 360\textdegree{} visual world creation with unified editing control. Code, model, and data will be released at our \href{https://world-shaper-project.github.io/}{project page}.

\end{abstract}   
\section{Introduction}
\label{sec:intro}

Unlike a perspective image, which depicts only a limited field of view from one direction, a panorama captures the \textit{entire visual world} surrounding the observer, offering a complete 360\textdegree{} perception of the scene. This makes it indispensable for applications such as immersive media~\cite{tukur2025panoramicImagingXR,lin2025surveyPerspectiveToPanorama}, 
environmental lighting~\cite{wang2022stylelightHDRPano,cheng2024evaluatingPano3DEstimation}, 
VR/AR~\cite{wang2025surveyTextDriven360Pano,bai2023localToGlobalPanoInpainting}, and autonomous driving~\cite{chugunov2024neuralLightSpheres,lin2025surveyPerspectiveToPanorama}.
Being able to generate and edit such panoramic worlds is therefore essential for creating and controlling realistic 360\textdegree{} experiences. 
However, editing in this 360\textdegree{} domain remains challenging.

\begin{figure}[t!]
    \centering
    \includegraphics[width=1\linewidth]{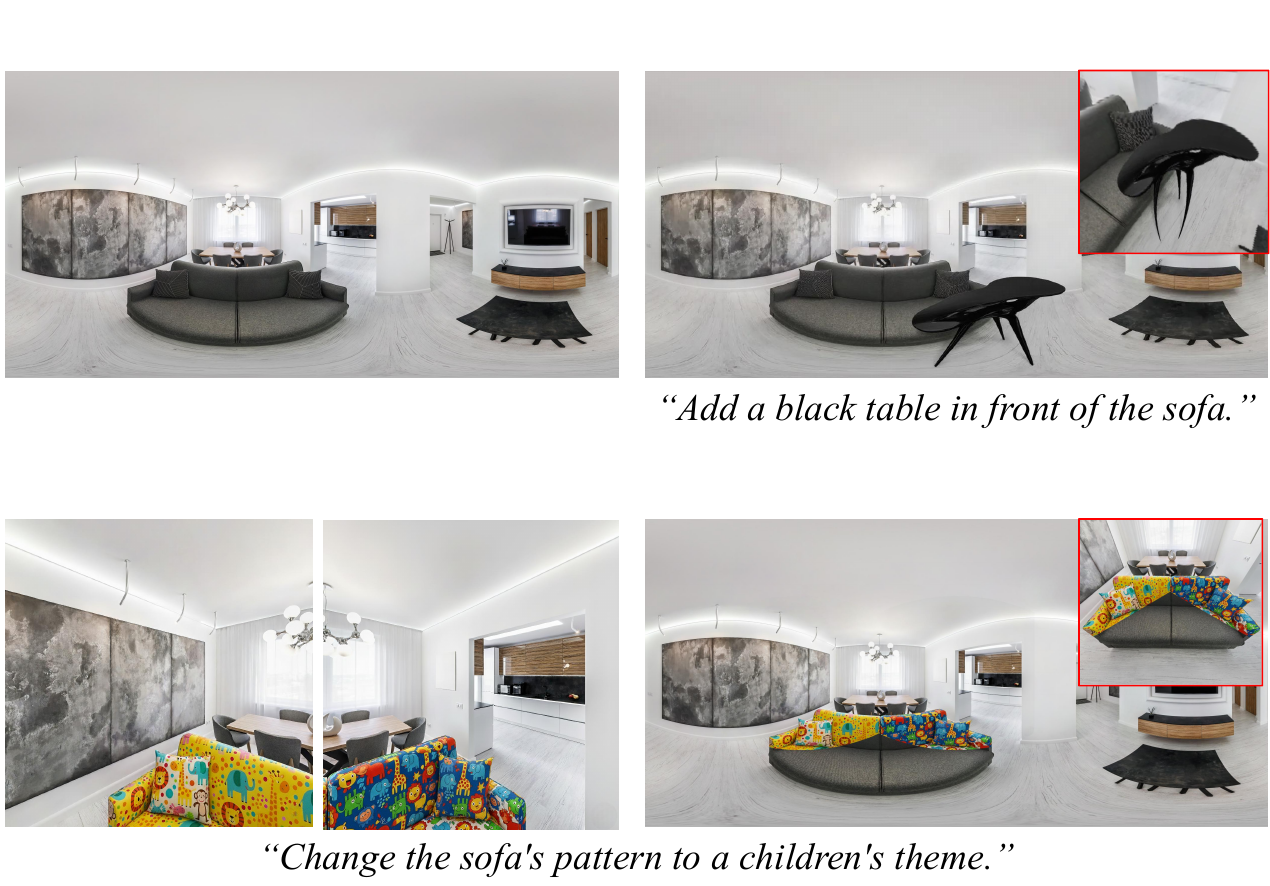}
    \put(-215,155){\scriptsize (a) Input Panoramic Image}
    \put(-110,155){\scriptsize (b) Nano Banana Pro~\cite{Sharon2025NanoBanana}}
    \put(-213,73){\scriptsize (c) Cube-maps after editing}
    \put(-100,73){\scriptsize (d) Reprojection back to ERP}
    \vspace{-2mm}
    \caption{Two paradigms for editing panoramic images using existing perspective-based models.
First row: applying Nano Banana Pro~\cite{Sharon2025NanoBanana} directly to ERP panoramas suffers from severe spatial distortion.
Second row: locally editing cube-map faces and reprojecting them back to ERP reduces distortion but breaks global coherence, causing discontinuities across cube boundaries. Refer to Sec.~\ref{sec:more_motivation} of Supplemental for more discussions.}
    \label{fig:observation}
    \vspace{-4mm}
\end{figure}
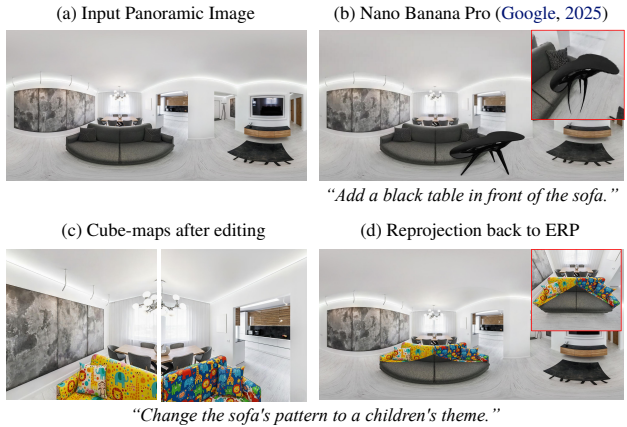

Despite the rapid progress in diffusion-based image generation~\cite{rombach2022high,flux2024,peebles2023scalable,achiam2023gpt} and editing~\cite{brooks2023instructpix2pix,batifol2025flux,wang2025gpt,zhang2025context}, these models are fundamentally designed to handle perspective images. 
Panoramic images, on the contrary, are commonly represented in the equirectangular projection (ERP)~\cite{lin2025one}, which maps the spherical environment to a 2D plane using latitude–longitude coordinates.
When applied directly to the ERP-based panoramas, perspective-based models struggle to deal with the spatial distortion, as shown in \figref{fig:observation}(b).
%
A trivial workaround is to convert the panorama into multiple cube-map faces, process each face independently, and then stitch them back together.
%
Such a decomposition, however, inevitably breaks global coherence: a single object may span multiple faces, as shown in \figref{fig:observation}(c) causing geometry and texture discontinuities when the cube faces are reassembled, as shown in \figref{fig:observation}(d).

Based on the above observation, we explore in this work an ERP-native formulation for panoramic editing, which offers two key advantages: 
(1) \textbf{\textit{Global Consistency}} -- the model perceives the entire 360\textdegree{} environment at once, enforcing semantic and geometric continuity across all directions and eliminating cross-face seams; and  
(2) \textbf{\textit{Unified Representation}} -- a single, continuous coordinate system aligns naturally with pre-trained diffusion models~\cite{wu2025qwenimagetechnicalreport}, forming an editing-centric representation where text prompts and spatial conditions apply consistently.

However, realizing this goal is far from trivial, due to the uneven geometric distortion in ERP images and the lack of paired data for panoramic editing.
To address these two challenges, we propose \name, a unified geometry-aware framework for direct panoramic image editing. 
We adopt a \textit{generate-then-edit} paradigm to overcome the lack of paired data for training editing models -- a controllable generation model is first trained to synthesize new objects conditioned on external constraints, which is then used to produce paired data for learning editing behaviors.

To handle geometric distortions, we introduce a \textit{geometry-aware learning strategy} that combines two principles: (i) \textbf{\textit{a position-aware shape constraint}}, which leverages feature- and output-level shape cues to impose spatially adaptive geometric supervision and encourage distortion-aware reasoning across different latitudes; and (ii) \textbf{\textit{a progressive curriculum training}}, which gradually shifts training from global panorama generation to localized object manipulation, allowing the model to internalize panoramic distortion priors.
Through this unified design, \name ~enables panorama generations and diverse editing capabilities such as object addition, removal, relocation, and appearance modification.

To verify the effectiveness of our \name, we curate a new panoramic editing benchmark, \dataset, and conduct extensive experiments across 
panorama object-level editing and generation.
\name ~demonstrates superior geometric consistency, editing fidelity, and text controllability on both synthetic and real-world panoramas. Beyond quantitative results, our framework empowers intuitive and flexible 360\textdegree{} visual world creation and editing. 
Furthermore, our framework can be progressively extended toward 3D world generation, enabling scene expansion and exploration from user-provided text or images.

Our main contributions can be summarized as follows:
\begin{itemize}
    \vspace{-3mm}
    \item 
    We introduce \name, a unified geometry-aware framework that enables diverse 360\textdegree{} editing within a single ERP-native representation, ensuring global consistency and seamless cross-view editing.
    \vspace{-3mm}
    \item 
    We propose two key mechanisms: \textit{a generate-then-edit pipeline} to synthesize paired panoramic data for supervised learning, and \textit{a geometry-aware learning strategy} to progressively align spatial distortions. 
    \vspace{-3mm}
    \item 
    We curate \dataset, a comprehensive panoramic editing benchmark, and conduct extensive experiments to demonstrate the superiority of our method.
\end{itemize}
\section{Related Works}
\label{sec:related_works}

\subsection{Text-to-Image Generation}
Recent advances in diffusion model~\cite{ho2020denoising, song2020denoising} have fundamentally reshaped text-to-image (T2I) generation. Early diffusion-based frameworks such as DALL{\textperiodcentered}E~\cite{ramesh2021zero,ramesh2022hierarchical,betker2023improving}, Imagen~\cite{saharia2022photorealistic}, and Stable Diffusion~\cite{rombach2022high, podell2023sdxl} established the foundation for scalable and high-quality image synthesis conditioned on language descriptions. 
More recently, DiT-based architectures have become the dominant paradigm. Representative models including SD3~\cite{stabilityai2024stable}, FLUX.1~\cite{flux2024}, Qwen-Image~\cite{wu2025qwenimagetechnicalreport}, and Seedream~\cite{Seedream2025v4} have demonstrated strong generative quality across diverse domains.
Yet text-only conditioning provides limited spatial and structural control, which has led to an increasing interest in controllable T2I frameworks~\cite{Zhang_2023_ICCV,Li_2024_ECCV_ControlNetPP,mou2023t2i,zhao2023unicontrolnet,Zhang_2025_ICCV_EasyControl,tan2025ominicontrol} to leverage additional modalities. 
Building on this, we adopt a controllable panoramic generation as an auxiliary stage of our generate-then-edit pipeline, enabling the construction of paired panoramic data for editing model training.

\subsection{Instruction-based Image Editing}

Image editing aims to find an optimal balance between image reconstruction and re-generation. Early approaches~\cite{hertz2022prompt,chefer2023attend,brooks2023instructpix2pix,gal2022image2stylegan,roich2022pivotal,tumanyan2023plugandplay} primarily operate in a training-free manner, relying on attention manipulation~\cite{hertz2022prompt,chefer2023attend}, prompt engineering~\cite{brooks2023instructpix2pix}, or latent-space inversion~\cite{mokady2023null,cao2023masactrl} to steer T2I models toward the desired edit.
To overcome the weak robustness, subsequent works~\cite{xiao2025omnigen,wu2025omnigen2,zhang2025context,batifol2025flux1kontext,wu2025qwenimage,sheynin2023emu} adopt a data-driven paradigm: they first construct large-scale editing datasets~\cite{Zhao2024UltraEdit, Yu2025AnyEdit, Ye2025ImgEdit}, and then train unified models to jointly understand instructions and generate edited images.
More recently, methods like~\cite{wu2025qwenimage, liu2025step1x, achiam2023gpt,Sharon2025NanoBanana} incorporate multimodal understanding into diffusion backbones, equipping editing models with world-level reasoning and yielding more consistent, semantically aligned edits. Despite their success, these methods are all based on perspective images and fail to handle spatial distortions inherent in panoramic projections.

\vspace{-2mm}
\subsection{Panorama Generation and Editing}

\textit{Text-to-panorama} generation aims to synthesize panoramic images from textual descriptions. Existing methods can be broadly grouped according to how they address the geometric challenges of the ERP representation.

Distortion-aware methods~\cite{chen2022text2light,sun2025spherical,wu2024spherediffusion,quattrini2024merging} modify networks or feature sampling to compensate for the non-uniform stretching.
Projection-driven methods~\cite{ccapuk2025tandit,zhang2024taming,park2025spherediff,kalischek2025cubediff} mitigate distortions by converting panoramas into alternative views (\eg, cubemap) and learning to fuse multi-projection features. 
Continuity-oriented methods
~\cite{liu2024panofree,feng2023diffusion360,tang2023mvdiffusionenablingholisticmultiview,feng2025dit360} focus on maintaining horizontal cyclic consistency. Beyond text-based synthesis, \textit{image-to-panorama}~\cite{akimoto2022diverse,wu2023panodiffusion,zheng2025panorama,zhang2025top2pano,huang2025omnix,wang2023360,schwarz2025recipe,yuan2025camfreediff} generation further incorporates NFOV images as explicit local constraints. Further, \textit{3D}~\cite{worldgen2025ziyangxie,team2025hunyuanworld,yang2025matrix3d,huang2025dreamcube} and \textit{video-based}~\cite{li2024omnidrag,xia2025panowan,wang2024360dvd,dong2025panolora,yang2024layerpano3d} panorama generation methods extend the traditional panorama generation task toward dynamic or explorable 360° environments generation.

Panorama editing methods like SE360~\cite{SE360} and Omni2~\cite{yang2025omni} rely heavily on cubemap projections. Their dataset generates by cubemap-inpainting, which restricts its edits to simple object-level operations. In contrast, our method operates directly in the ERP domain, preserving spherical continuity during both data synthesis and model training, enabling globally consistent geometry and richer edit types.

\section{Method}

\begin{figure*}[t]
    \centering
    \begin{overpic}[width=1\textwidth]{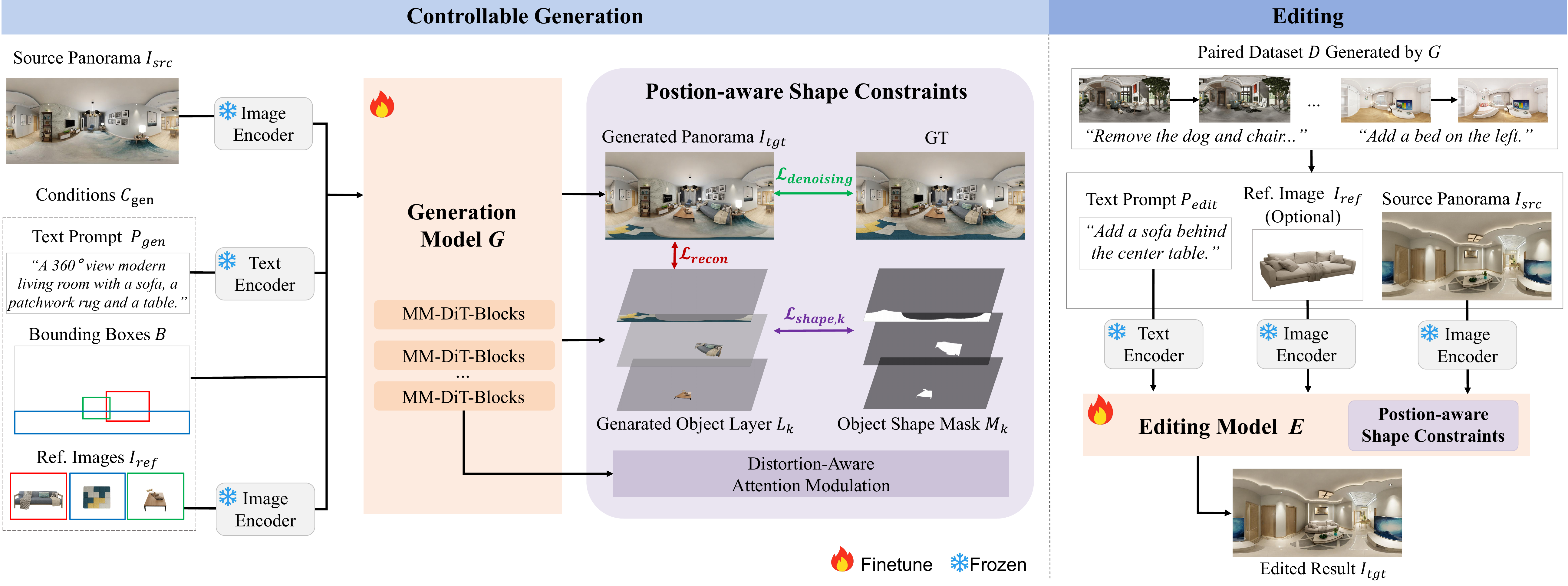}
    \end{overpic}
    \vspace{-6mm}
    \caption{Pipeline of \name. We adopt a \textbf{generate-then-edit} paradigm: we first train a controllable panoramic generator $G$ to synthesize diverse target panoramas $I_{tgt}$ from source panoramas $I_{src}$ under constraints $C_{gen}$. The synthesized pairs $(I_{src}, P_{\text{edit}}, I_{tgt})$ form a large-scale dataset $D$, on which an editing model $E$ is trained to perform instruction-driven panoramic editing. To address latitude-dependent geometric distortion in ERP images, we introduce a \textbf{geometry-aware learning} strategy that combines \textit{Progressive Curriculum Training}, which gradually shifting from global panorama generation to localized object manipulation, and \textit{Position-aware Shape Constraints}, which modulate distortion-aware attention and enforce layered shape consistency across object regions.}
    \label{fig:pipe}
\vspace{-3mm}
\end{figure*}

\subsection{Overview}
\label{subsec:overview}

Our goal is to develop a unified panoramic image editing framework, denoted as \name, which integrates controllable panoramic generation and editing within a single model. This design enables coherent 360° panoramic image synthesis and diverse editing operations, such as object addition, removal, relocation, and appearance modification. Formally, given a source panorama $I_{\text{src}}$ in the equirectangular projection (ERP) format and an editing instruction $C_{\text{gen}}$ (\eg, text prompt and/or spatial conditions), our model produces an edited panorama $I_{\text{tgt}}$ that satisfies the instruction while preserving global geometric consistency.

To achieve this, \emph{\name} integrates two complementary components. (1) \textit{Generate-then-Edit paradigm}. We first train a controllable panoramic generation model to synthesize diverse scenes under external constraints, which is then served as a data generator to construct paired data for supervised editing learning. (2) \textit{Geometry-Aware Learning}. To handle latitude-dependent ERP distortions, we introduce a geometry-aware design combining a position-aware shape constraint and progressive curriculum training, ensuring spatially adaptive supervision and distortion-aware reasoning.
\figref{fig:pipe} shows the overall architecture.

\subsection{Generate-then-Edit}
\label{subsec:data_gen}

To enable diverse panoramic editing, we first require a controllable generation model that can flexibly synthesize new content under user-defined conditions.

\vspace{-2mm}
\subsubsection{Controllable Panorama Generation}
An ideal generator should satisfy three criteria: (i) support multiple forms of control;  
(ii) handle multiple objects with coherent geometry; and (iii) produce distortion-consistent panoramas directly in the ERP domain.

Formally, we train a panoramic generation model $G$ that takes as input a source panorama $I_{\text{src}}$ and an external condition $C_{\text{gen}}$, and outputs a synthesized panorama $I_{\text{tgt}} = G(I_{\text{src}}, C_{\text{gen}})$, where the condition $C_{\text{gen}}$ provides both semantic and spatial guidance, and is defined as: $C_{\text{gen}} = \{ P_{\text{gen}}, \{(B_k, I_{\text{ref}}^k)\}_{k=1}^K \}$, where $P_{\text{gen}}$ is a text prompt describing the desired panoramic scene, and each tuple $(B_k, I_{\text{ref}}^k)$ denotes the $k$-th object’s spatial region specified by a bounding box $B_k$ and its corresponding reference image (optional). 
During inference, we encode the source panorama $I_{\text{src}}$, and each reference image $I_{\text{ref}}^k$ into latent features using the VAE encoder. Each bounding box $B_k$ is downsampled to match the latent resolution, and the resulting spatial masks are concatenated with the corresponding image and text tokens in a fixed order to form the conditioning input.
During training, we selectively drop elements from $C_{gen}$ to simulate diverse control scenarios. This stochastic conditioning encourages the model to adapt to varying levels of guidance, thereby enhancing its generalization and robustness.

\subsubsection{Paired Data Construction}
\label{sec:pair_data}

Build on the above model $G$, we construct a dataset 
$D = \{ (I_{\text{src}}, P_{edit}, I_{\text{tgt}})_i \}_{i=1}^N$, 
where each triplet consists of a source panorama $I_{\text{src}}$, an editing instruction $\mathcal{P}_{edit}$, and the edited target panorama $I_{\text{tgt}}$.
For each $I_{\text{src}}$, we synthesize multiple $I_{\text{tgt}}$ associated with distinct editing operations, covering five representative types:

\vspace{-5mm}
\begin{itemize}
    \item \textbf{Addition.}  We randomly sample several bounding boxes $B$ as potential insertion regions. For each $B$, GPT-5~\cite{ChatGPT} first generates a suitable object description; we then retrieve a matching reference image $I_{\text{ref}}$ from the internet and ask GPT-5 again to produce a global text prompt $P_{\text{gen}}$.
    Finally, the target panorama $I_{tgt}$ is synthesized via 
    $ G(I_{\text{src}}, \{ P_{\text{gen}}, \{B, I_{\text{ref}}\}\})$.
    \vspace{-3mm}
    \item \textbf{Removal.}  
    This is obtained by swapping the source ($I_{\text{src}}$) and target ($I_{tgt}$) images from the addition pairs.
    \vspace{-3mm}
    \item \textbf{Replacement.} It is generated by inserting two different objects at the same location $B$ within one $I_{\text{src}}$. The two generated panoramas form a replacement pair.
    \vspace{-3mm}
    \item \textbf{Movement.} It is produced by placing the same reference object $I_{\text{ref}}$ at different target locations $B_1$ and $B_2$.
    \vspace{-7mm}
    \item \textbf{Modification.} It is created by applying two addition operations at the same position using slightly different reference images $I_{\text{ref}}^{(1)}$ and $I_{\text{ref}}^{(2)}$, where $I_{\text{ref}}^{(2)}$ is obtained via the editing model~\cite{batifol2025flux1kontext}. For global changes, $G$ performs scene-level edits from text prompts.
\end{itemize}
\vspace{-3mm}
With constructed data, GPT-5 automatically compares each $(I_{\text{src}}, I_{\text{tgt}})$ pair and generates the corresponding instruction $P_{\text{edit}}$, resulting in a large-scale, diverse dataset for supervised training.
Finally, to further ensure data quality, trained annotators are asked to lightly validate a subset of pairs.

\subsection{Geometry-Aware Learning Strategy}
\label{subsec:geometry}

To ensure geometrically consistent generation and editing, 
we introduce two complementary designs:
a \emph{position-aware shape constraints} 
for spatially adaptive geometric supervision, and a \emph{progressive curriculum training} strategy that gradually shifts learning  from global panorama generation to localized object manipulation.
\label{subsec:geometry}
\subsubsection{Position-Aware Shape Constraints}

To \emph{explicitly} handle spatially varying distortions in ERP panoramas, we introduce feature- and output-level mechanisms to enforce geometry-aware shape consistency.

\vspace{1ex}
\noindent \textbf{Distortion-Aware Attention Modulation.}
At the feature level, we modulate object-specific attention maps for distortion-aware spatial focus.
Specifically, we first extract the object-relevant cross-attention map $A$ between image tokens $\mathbf{X}$ and the object text embeddings $\mathbf{E}_o$.  
It is then modulated using the ground-truth object mask $M$  with spatially adaptive strength.
We adopt a residual-based modulation~\cite{kim2023densetexttoimagegenerationattention}, where the modulation residual is computed as:
\vspace{-2mm}
\begin{equation}
    R = \bm{\alpha} \odot \left(M \odot (A_{\max} - A)- (1 - M) \odot (A - A_{\min})\right),
\end{equation}
where $\bm{\alpha}$ is a spatially adaptive modulation map determined by the latitude of each spatial location.
For each pixel (or image token) at vertical coordinate $y$, we define:
\begin{equation}
\bm{\alpha}(y)
= 1 - \cos\left( \frac{\pi}{2} - \frac{\pi y}{H} \right),
\label{eq:alpha_distortion}
\end{equation}
where $\bm{\alpha}(y)$ increases toward higher latitudes.
In this way, the attention map is finally updated as $A' = A + R$.

\vspace{1ex}
\noindent \textbf{Layered Shape Loss.} 
At the output level, we render objects into transparent layers and apply a layer-wise shape loss to ensure geometric consistency.
Specifically, besides the generated panorama image, the model additionally outputs
a set of $K$ transparent object layers $\{L_1, \ldots, L_K\}$, where each layer corresponds to one input object.
Each layer $L_k$ is represented as an RGBA image, whose alpha channel provides a direct estimate of the object’s shape mask $M_k$. 
In our implementation, since the employed VAE~\cite{wu2025qwenimage} Decoder produces only RGB outputs, we approximate the RGBA representation by generating RGB images with a pure-white background and then derive the object’s alpha mask $M_k$ via connectivity-based segmentation.
To supervise these layers, we impose two complementary constraints.
First, we use an IOU loss 
$\mathcal{L}_{\text{shape}, k} = \alpha_k \cdot (1 - \text{IoU}(M_k, M_{\text{shape}, k}))$ to ensure that the predicted mask $M_k$ aligns with its GT shape mask $M_{\text{shape}, k}$,  where $\alpha_k$ is a spatially adaptive coefficient computed based on Eq.~\ref{eq:alpha_distortion}.
Second, we introduce a reconstruction loss to enforce each predicted layer $L_k$ to be consistent with its appearance in the final panorama,
by reconstructing the composite image
$I_{\text{comp}} = \text{Composite}(I_{\text{src}}, L_1, \ldots, L_K)$  and matching it 
with $I_{\text{tgt}}$ within the object’s bounding box $B_k$, as:  

\vspace{-4mm}
\begin{equation}
    \mathcal{L}_{\text{recon}} = \frac{1}{N} \sum_{i=1}^{N} \| I_{\text{comp}}(i) - I_{\text{tgt}}(i) \|_2^2,    
\end{equation}
\vspace{-4mm}

where $N$ is the total number of pixels. 
The final layered shape loss aggregates both objectives across all $K$ objects:
\vspace{-3mm}
\begin{equation}
    \mathcal{L}_{\text{shape}} =  \frac{1}{K} \sum_{k=1}^{K}  \left( \mathcal{L}_{\text{shape}, k} \right) + \mathcal{L}_{\text{recon}}.
\end{equation}
\vspace{-6mm}

This layered supervision delivers explicit geometric guidance, prompting accurate and distortion-aware object shapes across the entire panorama.

\subsubsection{Progressive Curriculum Training}
\label{sec:training}

While the position-aware shape constraints  explicitly regulate geometric distortion, they remain limited by the diversity of training instances. 
In contrast, global panoramic generation tasks naturally expose the model to abundant and diverse distortion configurations. 
To leverage these rich priors, we 
gradually transitions learning from global panorama generation to localized object manipulation, enabling the model to internalize panoramic distortion priors \emph{implicitly}.

\vspace{1ex}
\noindent \textbf{Stage 1: Global Structure and Distortion Learning.}  
%
%
%
%
In this stage, the model focuses solely on global panorama generation, aiming to capture the overall 360° scene structure and latitude-dependent distortion patterns. 
To this end, we train the model on both \emph{Text-to-Panorama} (T2P), which input is a textual description, and \emph{Image-to-Panorama} (I2P), which inputs include both a perspective image and a corresponding text prompt.
The standard denoising loss $\mathcal{L}_{\text{denoising}}$ is applied to reconstruct the panorama.

\vspace{1ex}
\noindent \textbf{Stage 2: Localized Controllable Generation.}  
In this stage, the model is transitioned to fine-grained,  object-level control. 
To achieve this, we train
the model to synthesize new objects under explicit spatial (\eg, via bounding boxes) and appearance (\eg, via reference images) constraints. 
Specifically, given a source panorama $I_{\text{src}}$ and control signals   
$\mathcal{C}_{\text{gen}} = \{ P_{\text{gen}}, B, I_{\text{ref}}\}$,  
the model $G$ generates a new panorama $I_{\text{tgt}}$ with the specified objects inserted. 
During training, both the layered shape loss $\mathcal{L}_{\text{shape}}$ and  the denoising loss $\mathcal{L}_{\text{denoising}}$ are applied.
The resulting controllable generation model is used as a data generator 
for Stage~3.

\vspace{1ex}
\noindent \textbf{Stage 3: Supervised Editing Learning.} 
Building on the controllable generation model $G$ from Stage~2 and paired data construction in \secref{subsec:data_gen}, we synthesize triplets $(I_{\text{src}}, \mathcal{P}_{edit}, I_{\text{tgt}})$ and 

train the final editing model $E$ on these instruction-driven pairs.
During training, we apply both layered shape loss $\mathcal{L}_{\text{shape}}$ and the denoising loss $\mathcal{L}_{\text{denoising}}$.

\vspace{1ex}
\noindent \textbf{Progressive Task Transition.}  
To prevent instability from abrupt task switching, we gradually transition training across stages. 
Data from the previous stage is linearly decayed as new-stage data increases, while a small portion of earlier data is retained to preserve prior capabilities. 
This strategy enables continual learning without forgetting, and yields a unified model covering all tasks.

\section{Experiments}

\subsection{Evaluation Protocol}
\label{subsec:evaluation}

\subsubsection{Setup}
\label{subsubsec:setup}

We adopt Qwen-Image-Edit-2509~\cite{wu2025qwenimagetechnicalreport} as the base model for panorama generation and editing, and fine-tune it using LoRA~\cite{hu2022lora} with rank 32, applied to the attention layers. 
By default, all panoramic images are trained at a resolution of $512 \times 1024$. 
All stages are trained on $8$ NVIDIA A100 GPUs with a per-GPU batch size of $4$, using AdamW with a learning rate of $1\times10^{-4}$.
Stage 1 is trained for 7k steps (8 hours), while stages 2 and 3 are each trained for 10k steps, taking approximately 22 hours and 18 hours, respectively. 
Training data for stages 1 and 2 is sourced from Sun360~\cite{xiao2012recognizing}, Structured3D~\cite{zheng2020structured3d}, and Pano360~\cite{SPEC:ICCV:2021}, along with additional images rendered from UE scenes. The UE scenes are also used to render the paired training data for stage 3.
To enable high-resolution editing (\eg, 2K), we discard the layering-related input/output components and fine-tune the model for 1500 additional iterations.

\subsubsection{Dataset}
\label{subsubsec:dataset}

We present \textbf{\dataset}, a comprehensive panorama editing benchmark that includes both real-world and synthetic panoramas, covering indoor and outdoor scenes.
\textbf{\dataset} consists of 200 panoramas (100 real-world and 100 synthetic) designed to evaluate the editing performance across diverse environments.
\vspace{-3mm}
\begin{itemize}
    \item \textit{Real-World Data:} 
    We have collected 100 panoramas from online sources (\eg, Freepik). To ensure diversity, GPT-5~\cite{ChatGPT} is used to generate 20 distinct search queries (10 indoor and 10 outdoor). For each query, five high-quality images are manually curated.

    \item \textit{Synthetic Data:} 
    We have generated 100 panoramas from two complementary sources: (i) Skybox~\cite{skybox}, an AI-based 360$^{\circ}$ content generator using diverse  prompts; (ii) Unreal Engine~\cite{unrealengine}, rendering panoramas from 3D assets.
\end{itemize}
\vspace{-3mm}

To evaluate panorama editing, each test case consists of a $(\text{source image}, \text{editing prompt})$ pair.
For each panorama, we define five editing tasks: \emph{removal}, \emph{addition}, \emph{replacement}, \emph{movement}, and \emph{modification}.
We employ GPT-5 to automatically detect objects  and generate  task-specific editing instructions conditioned on the identified object categories, yielding 1{,}000 test pairs (200 images $\times$ 5 editing tasks). Refer to \secref{supp:dataset} of Supplemental for more details.

\subsubsection{Metrics}
\label{subsubsec:metric}

We evaluate our framework using a distinct set of metrics for the panorama editing tasks.
Following Step1X-Edit~\cite{liu2025step1x}, we adopt the comprehensive \textit{VIEScore}~\cite{ku2024viescore} to evaluate editing quality from three perspectives:
\textit{(1) SC (Semantic Consistency)}: measuring how well the edited image aligns with the given instruction. 
\textit{(2) PQ (Perceptual Quality)}: assessing visual realism and the presence of artifacts. 
\textit{(3) O (Overall Score)}: combining the above two aspects as a  unified score.
Additionally, we also introduce the
\textit{Fréchet Inception Distance (FID)}~\cite{heusel2017gans} to evaluate the overall visual quality and \textit{
CLIP Direction Score~\cite{shi2024seededit}
}
to measure consistency between the edited image and the instruction.

\subsection{Comparisons with SOTAs}

We compare \name~with 8 SOTA editing methods: Qwen-Image-Edit-2509~\cite{wu2025qwenimagetechnicalreport}, FLUX.1 Kontext [dev]~\cite{batifol2025flux}, GPT-5~\cite{ChatGPT}, Nano Banana Pro~\cite{Sharon2025NanoBanana}, OmniGen2~\cite{wu2025omnigen2}, Omni\^2~\cite{omni2}, Step1X-Edit~\cite{liu2025step1x}, and IC-Edit~\cite{zhang2025context}. For fairness, we retrain all methods with available training codes.

\begin{table*}[t!]
\centering
\small
\caption{Quantitative comparison of our method against eight SOTA image editing methods on \textbf{panorama editing}.
The user study reports the average ranking
in image quality (IQ), distortion accuracy (DA), and text alignment (TA).
The best and second-best results are highlighted in \textbf{bold} and \underline{underlined}, respectively.
$^{\dag}$ denotes proprietary methods which cannot be re-trained.
All CLIP$_{dir}$ scores are multiplied by 100 for readability.}
\vspace{-2mm}
\setlength{\tabcolsep}{3pt}
\renewcommand{\arraystretch}{1.1}
\scalebox{0.68}{
\begin{tabular}{l|l|ccccccccc}
\toprule
\multirow{1}{*}{\textbf{Category}} & \multirow{1}{*}{\textbf{Metrics}}
& \makecell{Qwen\\~\cite{wu2025qwenimagetechnicalreport}}
& \makecell{Kontext\\~\cite{batifol2025flux}}
& \makecell{GPT-5$^{\dag}~$\\~\cite{ChatGPT}}
& \makecell{Nano Banana Pro$^{\dag}$\\~\cite{Sharon2025NanoBanana}}
& \makecell{IC-Edit\\~\cite{zhang2025context}}
& \makecell{Omni$^2$\\~\cite{omni2}}
& \makecell{Step1X-Edit\\~\cite{liu2025step1x}}
& \makecell{OmniGen2\\~\cite{wu2025omnigen2}}
& Ours \\
\midrule
\multirow{5}{*}{Automatic}
& FID $\downarrow$ & 65.42 & 60.18 & 92.37 & \underline{51.10} & 84.83 & 61.15 & 78.69 & 66.53 & \textbf{45.25} \\
& $\text{CLIP}_{\text{dir}}$ $\uparrow$ & 14.57 & 13.05 & 12.92 & \underline{16.74} & 16.21 & 16.65 & 11.93 & 13.24 & \textbf{19.62} \\
& SC $\uparrow$ & 5.91 & 3.62 & 2.53 & 5.88 & 5.74 & 5.86 & 5.41 & \underline{5.98} & \textbf{8.15} \\
& PQ $\uparrow$ & 5.42 & 4.03 & 4.91 & 5.41 & 5.23 & 5.36 & \underline{5.48} & 5.11 & \textbf{7.87} \\
& O $\uparrow$ & \underline{5.66} & 2.77 & 2.73 & 5.64 & 5.48 & 5.60 & 5.44 & 5.53 & \textbf{8.01} \\
\midrule
\multirow{3}{*}{User study}
& IQ $\downarrow$ & 4.58 & 7.23 & 8.91 & \underline{2.37} & 6.76 & 5.05 & 5.77 & 3.14 & \textbf{1.19} \\
& DA $\downarrow$ & \underline{2.83} & 6.94 & 8.78 & 3.41 & 7.07 & 5.22 & 5.34 & 4.15 & \textbf{1.26} \\
& TA $\downarrow$ & 3.31 & 7.52 & 8.83 & \underline{2.18} & 4.95 & 5.09 & 8.02 & 3.76 & \textbf{1.34} \\
\bottomrule
\end{tabular}}
\label{tab:sota}
\vspace{0mm}
\end{table*}

\vspace{-3mm}
\begin{figure*}[t!]
    \centering
    \footnotesize
    \includegraphics[width=1\linewidth]{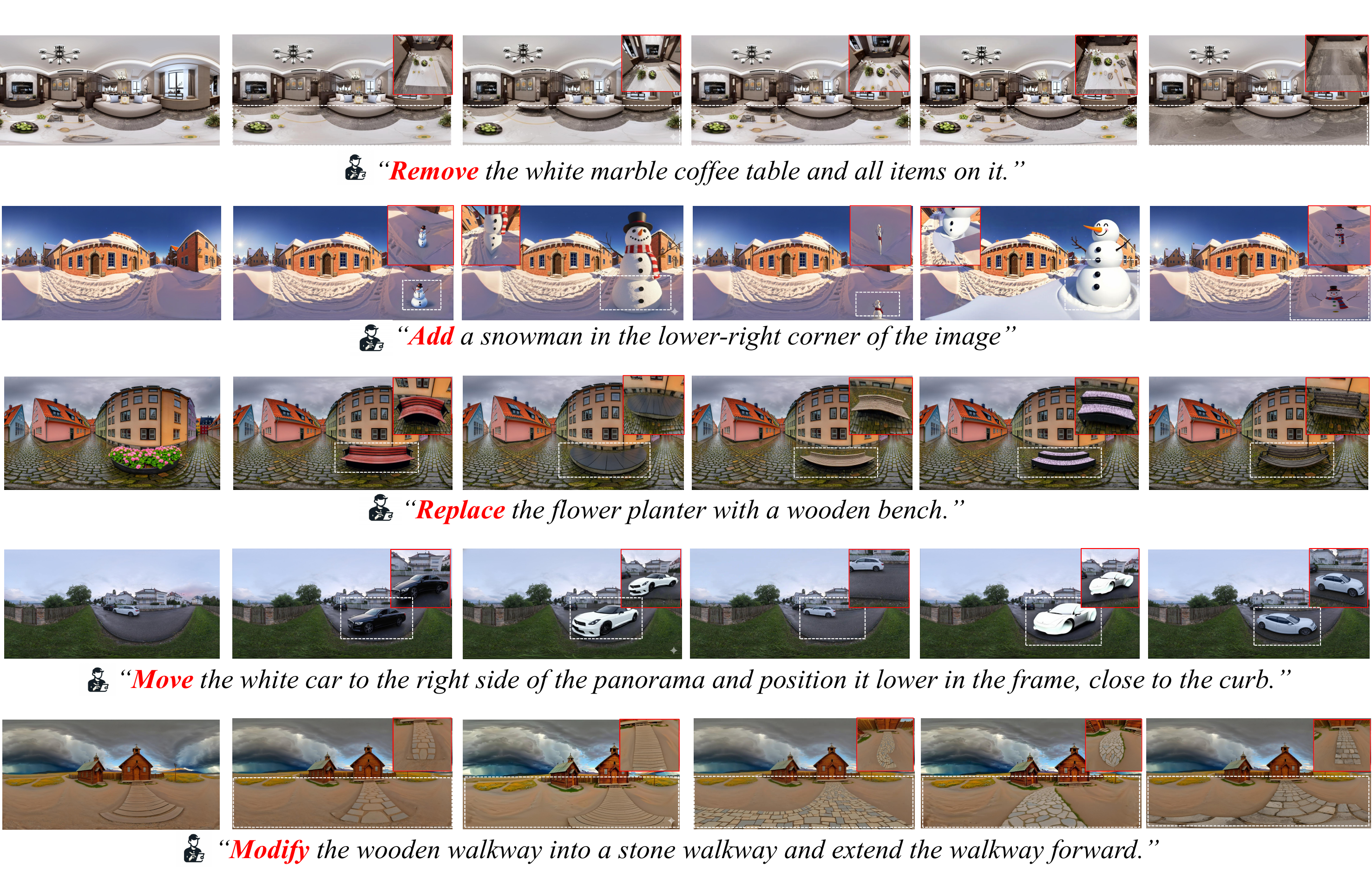}
    \put(-475,308){\small{(a) Input Image}}
    \put(-730,308){\small{\makecell{(b) Qwen\\~\cite{wu2025qwenimagetechnicalreport}}}}
    \put(-570,308){\small{\makecell{(c) Nano Banana Pro\\~\cite{Sharon2025NanoBanana}}}}
    \put(-410,308){\small{\makecell{(d) Omni2\\~\cite{omni2}}}}
    \put(-245,308){\small{\makecell{(e) OmniGen2\\~\cite{wu2025omnigen2}}}}
    \put(-80,308){\small{\makecell{(f) Ours}}}
    \vspace{-2mm}
    \caption{Qualitative comparison with four top-performing SOTA methods from Tab.~\ref{tab:sota} on panorama editing. White boxes indicate the selected regions for visualization. The corresponding edited areas are shown in the perspective view within the red boxes. Refer to Sec.~\ref{sec:more_results} of Supplemental for more results.} 
    \vspace{-3mm}
    \label{fig:sota_qual}
\end{figure*}
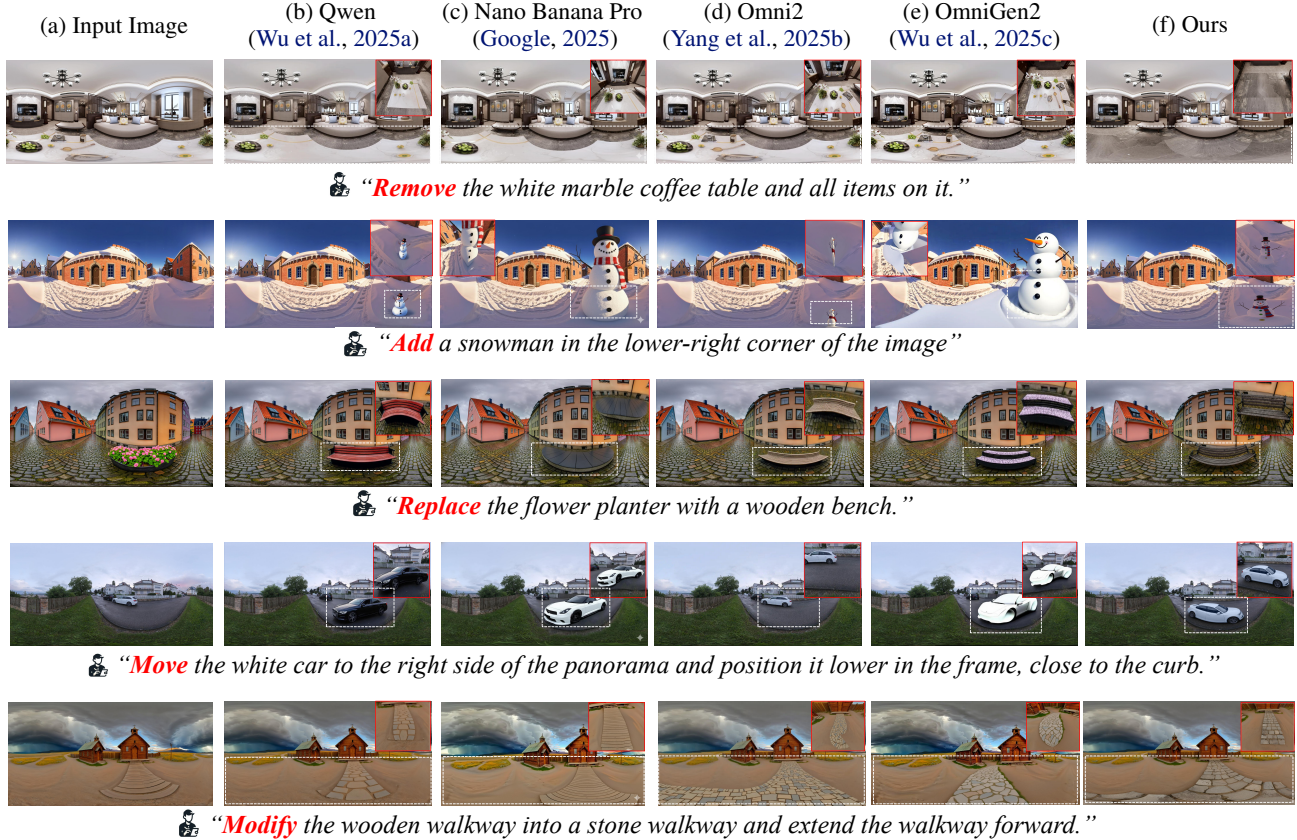

\vspace{2mm}
\noindent \textbf{Quantitative Comparison.} 
    Tab.~\ref{tab:sota} presents the quantitative comparison.
    Our method achieves the best performance on the Overall (O) metric and consistently surpasses all competing approaches across every criterion, demonstrating its comprehensive advantages.
    Specifically, it attains the lowest FID and highest PQ, indicating  superior visual realism and perceptual quality of the generated results.
    Moreover, the highest $\text{CLIP}_{\text{dir}}$ and SC further show that our model
    most faithfully adheres to the editing instructions.

\noindent \textbf{Qualitative Comparison.} 
\label{sec:qual_comp}
Fig.~\ref{fig:sota_qual} shows the visual comparison. 
Existing methods exhibit two major limitations. 
First, \textbf{Distortion Unawareness} — they fail to perceive or reason about geometric distortions in panoramic images, leading to various failures depending on the editing type. 
For \textit{removal} (row 1), distorted objects are misrecognized or ignored (\eg, only part of the objects are removed because the distorted one is not correctly identified). 
For \textit{addition} and \textit{movement} (rows 2 and 4), newly generated or repositioned objects exhibit unnatural shapes or inconsistent geometry due to incorrect distortion handling. 
Second, \textbf{Background Instability} — some methods, such as Nano Banana Pro and OmniGen2, tend to modify the background regions that should remain unchanged ((c) and (e), row 4). 
In contrast, our method effectively mitigates both issues by maintaining geometric consistency and precise text-guided control. Refer to Sec.~\ref{sec:prompt_enhance} of Supplemental for more discussions.

\vspace{-2mm}
\begin{figure*}[t!]
    \centering
\includegraphics[width=1.0\linewidth]{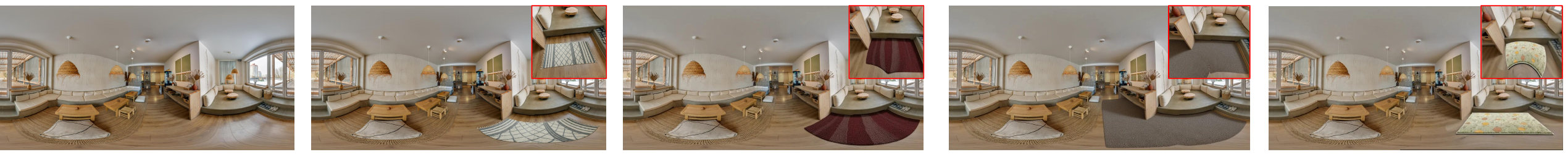}
    \put(-468,51){\small{(a) Input Image}}
    \put(-366,51){\small{(b) with all}}
    \put(-272,51){\small{(c) w/o LSL}}
    \put(-175,51){\small{(d) w/o DAAM}}
    \put(-86,51){\small{(e) w/o LSL+DAAM}}
    \vspace{-3mm}
    \caption{Qualitative ablation of Position-Aware Shape Constraints. Editing prompt: ``\emph{\textbf{Add} a carpet to the right side.}''}

\label{fig:ablation_loss}
\end{figure*}

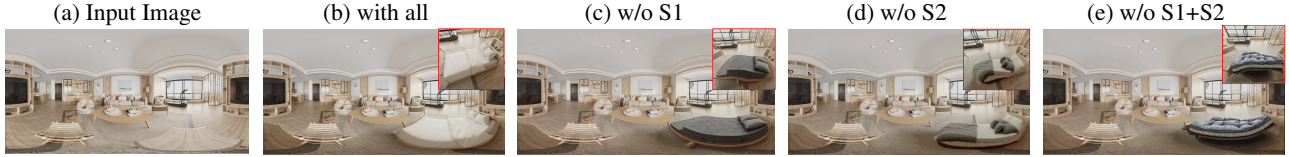
\begin{figure*}[t]
    \centering
\includegraphics[width=1.0\linewidth]{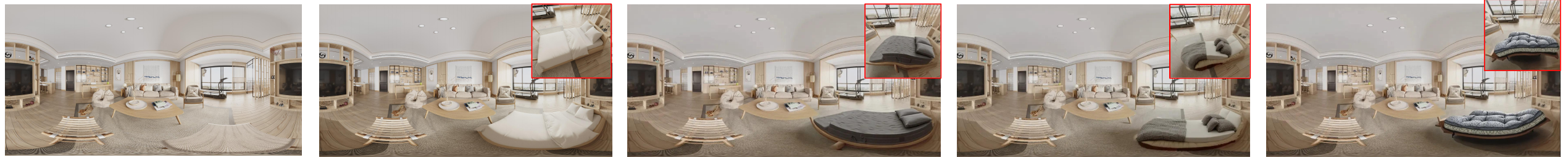}
    \put(-468,51){\small{(a) Input Image}}
    \put(-366,51){\small{(b) with all}}
    \put(-267,51){\small{(c) w/o S1}}
    \put(-168,51){\small{(d) w/o S2}}
    \put(-77,51){\small{(e) w/o S1+S2}}
    \vspace{-2mm}
    \caption{Qualitative ablation of Progressive Curriculum Training. Editing prompt: ``\emph{\textbf{Replace} the wooden stool on the right with a bed.}''}
    \label{fig:ablation_stage}
\end{figure*}

\noindent{\textbf{User Study.}}
We also conduct a user study with $72$  participants to evaluate human preferences.
For the editing task, each participant is shown 20 cases.
Each case contains (1) an input panorama,  (2) an editing instruction, and (3) nine randomly shuffled results from \name ~and the eight competing methods. 
Participants rank all methods according to three criteria: image quality (IQ), distortion accuracy (DA), and text alignment (TA). 
We aggregate rankings over all participants and report the averaged scores
in the bottom part of \tableref{tab:sota}. These results show that our method is consistently ranked highest on all three aspects.

\subsection{Ablation Study}
\label{subsubsec:ablation}

\begin{table}[ht!]
\centering
\caption{Ablation study on removing key components of the position-aware shape constraints:
\textit{Distortion-aware Attention Manipulation} (DAAM) and \textit{Layered Shape Loss} (LSL).
}
\vspace{-2mm}
\label{tab:ablation_components}
\resizebox{\linewidth}{!}{%
\begin{tabular}{cc|ccccc}
    \toprule
    DAAM & LSL & \textbf{PQ} $\uparrow$  & \textbf{SC} $\uparrow$  & \textbf{O} $\uparrow$  & \textbf{FID} $\downarrow$  & \textbf{$\text{CLIP}_{\text{dir}}$} $\uparrow$ \\
    \midrule
     &   & 5.65 & 5.02 & 5.32 & 58.32 & 16.36 \\
     \checkmark &   & 6.92 & 7.84 & 7.45 & 53.82 & 18.04 \\
    &  \checkmark   & 6.85 & 7.76 & 7.38 & 54.47 & 17.85 \\
    \checkmark & \checkmark  & \textbf{7.87} & \textbf{8.15} & \textbf{8.01} & \textbf{45.25} & \textbf{19.62}  \\
    \bottomrule
\end{tabular}%
}
\vspace{-2mm}
\end{table}

\noindent \textbf{Position-Aware Shape Constraints.} 
In this ablation,  the progressive curriculum training is enabled by default.
Since both Stage~2 and Stage~3 employ shape constraints, we ablate them in \tableref{tab:ablation_components} by removing (i) \emph{both} constraints simultaneously (first-row), (ii) the \emph{output-level} (layered shape) constraint from \emph{both} stages (second-row), or (iii) the \emph{feature-level} constraint from \emph{both} stages (third-row).

\figref{fig:ablation_loss} shows the visual result. We draw three main conclusions.
\ding{172} Using only the denoising loss leads to the poorest performance (\figref{fig:ablation_loss}(e)), especially in terms of semantic consistency (SC) and overall visual quality (FID).
\ding{173} Removing either layered shape loss (LSL) or distortion-aware attention modulation (DAAM) leads to noticeable performance deterioration, obvious geometric distortions (\figref{fig:ablation_loss}(c)) or visual artifacts (\figref{fig:ablation_loss}(d)), respectively.
\ding{174} Combining both components yields complementary benefits, enabling our approach to achieve the best geometric stability and visual quality (\figref{fig:ablation_loss}(b)).

\noindent \textbf{Progressive Curriculum Training.}
In this ablation, Position-aware shape constraints are enabled by default. 
We evaluate the editing training (Stage~3) under three settings by removing earlier stages: 
(i) remove \emph{both} Stage~1 and Stage~2 (\ie, train Stage~3 directly from the Qwen-Image-Edit weights);
(ii) remove \emph{only} Stage~2 (\ie, keep only the global generation weights from Stage~1); and 
(iii) remove \emph{only} Stage~1 (\ie, keep only the localized controllable generation weights from Stage~2).

\figref{fig:ablation_stage} and ~\tableref{tab:ablation_progressive} show the results. We draw three conclusions.
\ding{172} Removing Stage~1 (\tableref{tab:ablation_progressive}(row 3)) weakens global structural priors, degrading overall geometry and generative metrics (\eg, FID). 
\ding{173} Removing Stage~2 (\tableref{tab:ablation_progressive}(row 2)) harms fine-grained controllability, lowering editing-related metrics (\eg, PQ and $\text{CLIP}_{\text{dir}}$).
\ding{174} Eliminating \emph{both} stages (\tableref{tab:ablation_progressive}(row 1)) leads to the largest drop across all metrics and incurs the largest distortion, while keeping \emph{both} (\tableref{tab:ablation_progressive}(row 4)) achieves the best performance, indicating that Stage~1 and Stage~2 are complementary.

\begin{table}[ht!]
\centering
\caption{Ablation study on removing key stages of the progressive curriculum training:
\textit{Global Structure and Distortion Learning} (S1) and \textit{Localized Controllable Generation} (S2). 
}
\vspace{-2mm}
\label{tab:ablation_progressive}
\resizebox{0.9\linewidth}{!}{
\begin{tabular}{cc|ccccc}
    \toprule
    S1 & S2 & \textbf{PQ} $\uparrow$  & \textbf{SC} $\uparrow$  & \textbf{O} $\uparrow$  & \textbf{FID} $\downarrow$  & \textbf{$\text{CLIP}_{\text{dir}}$} $\uparrow$ \\
    \midrule
    &  & 6.74 & 7.31 & 7.18 & 56.47 & 17.74 \\
    \checkmark &  & 7.25 & 7.76 & 7.50 & 48.91 & 18.36 \\
    & \checkmark & 7.02 & 7.68 & 7.43 & 52.33 & 19.11\\
    \checkmark & \checkmark & \textbf{7.87} & \textbf{8.15} & \textbf{8.01} & \textbf{45.25} & \textbf{19.62} \\
    \bottomrule
\end{tabular}}
\vspace{-3mm}
\end{table}

\section{Conclusion}
In this paper, we have presented \name, a unified framework for panorama editing. 
We introduced a generate-then-edit paradigm for supervised training, along with a geometry-aware learning strategy for global consistent distortion generation.  
Extensive experiments on the \dataset~ show that our method achieves superior geometric fidelity, editing controllability, and visual realism. 

\begin{figure}[ht!]
    \centering
\includegraphics[width=1.0\linewidth]{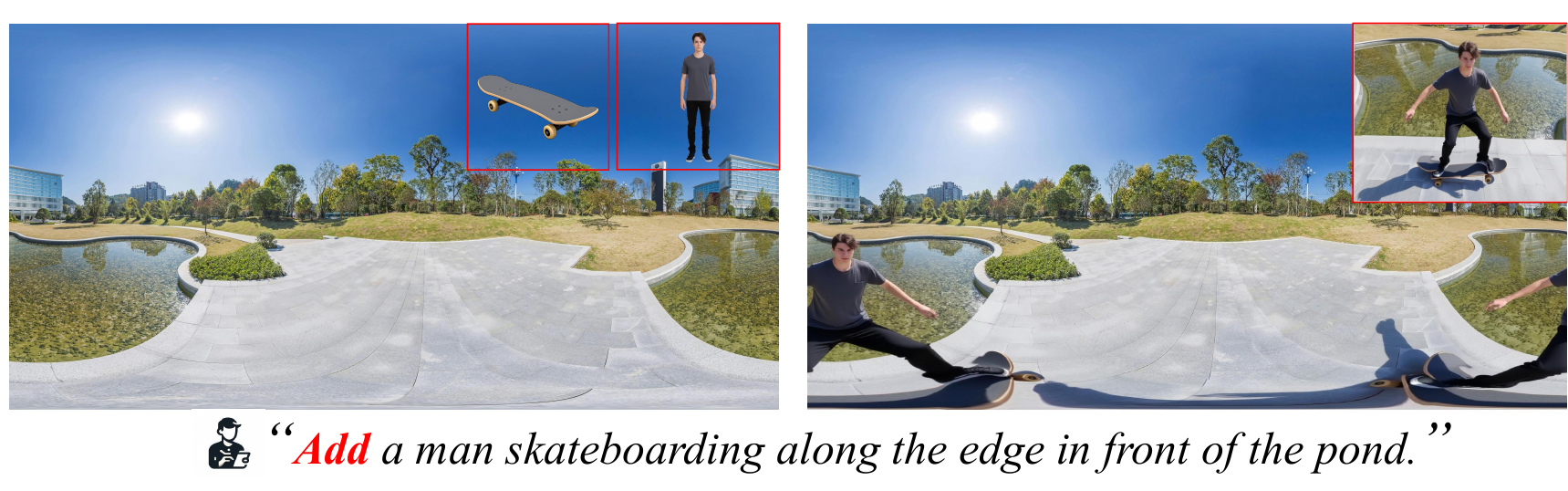}
    \put(-240,75){\small{(a) Input Panorama \& Ref. Images}}
    \put(-72,75){\small{(b) Ours}}
    \vspace{-3mm}
    \caption{An example failure case of \name.}
    \label{fig:limitation}
    \vspace{-3mm}
\end{figure}

Our work has limitations. 
When handling object–object interactions, artifacts may appear around the interaction region, as shown in \figref{fig:limitation}. 
Future work will incorporate explicit inter-object modeling to enhance physical coherence.

\section*{Impact Statement}

This paper presents work whose goal is to advance the field of Machine Learning. There are many potential societal consequences of our work, none which we feel must be specifically highlighted here.

\bibliography{main}
\bibliographystyle{icml2026}

\newpage
\clearpage
\appendix

\twocolumn[
\centering
{\Large \bfseries World-Shaper: A Unified Framework for 360\textdegree{} Panoramic Editing\par}
\vspace{4pt}
{\Large Appendix\par}
\vspace{1em}
]

\section{Overview}

In this appendix, we provide additional implementation details, ablation studies, and extended evaluations that further support and expand the findings presented in the main paper. We include the following components:

\begin{itemize}
    \vspace{-2mm}
    \item \textbf{Dataset construction and statistics.} Statistical analyses of the \dataset\ and detailed descriptions of the dataset construction procedures used for training our controllable generation and editing models (Sec.~\ref{supp:dataset}).
    \vspace{-2mm}
    \item \textbf{Model implementation.} Extended implementation details, including full network configurations and the specialized training techniques used in practice (Sec.~\ref{supp:model}).
    \vspace{-6mm}
    \item \textbf{Additional experiments.} Further ablations and extended comparisons for both editing and generation tasks (Sec.~\ref{supp:exp}).
    \vspace{-2mm}
    \item \textbf{Application details.} Diverse applications enabled by our method (Sec.~\ref{supp:application}).
\end{itemize}

\section{Dataset Construction \& Analysis}
\label{supp:dataset}

Two distinct data components facilitate our work: (1) a large-scale Training Dataset, essential for the proposed progressive training curriculum, and (2) the challenging Test Benchmark, \dataset, necessary for rigorous validation and comparison. This section details the construction methodology and statistical analysis of both.

\subsection{\dataset~ Construction}

In the main paper, we briefly introduced our panorama editing benchmark \dataset. To comprehensively evaluate our method on panorama generation as well, we extend \dataset\ to cover both panorama editing and panorama generation tasks, including \textbf{text-to-panorama} and \textbf{image-to-panorama} generation. As shown in Figure~\ref{fig:construct_bench}, the expanded benchmark is constructed through the following steps:

\begin{figure*}[h]
\centering
\includegraphics[width=1.0\linewidth]{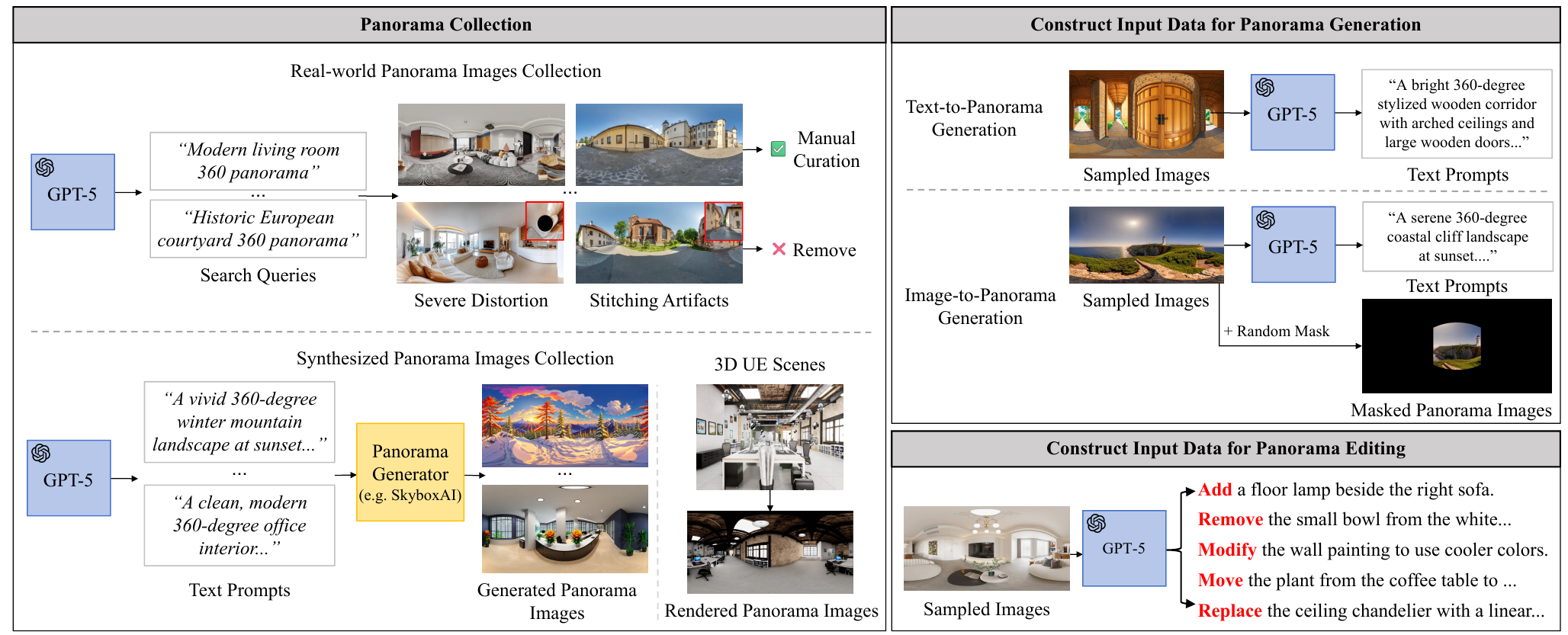}
\caption{Overview of the construction process of our \dataset.}
\label{fig:construct_bench}
\end{figure*}

\subsubsection{Panorama Collection}

In addition to the 200 high-quality panorama images collected for the panorama editing task, we employed the identical stringent collection standards to acquire a separate set of 200 panorama images dedicated solely to the evaluation of panorama generation methods.

\subsubsection{Construct Input Data for Panorama Generation}
We split the panorama generation task into two sub-tasks:
\begin{itemize}
\item \textit{Text-to-Panorama:} Each source image is first described by GPT-5~\cite{ChatGPT}, which produces a rich, scene-level caption serving as the input text prompt for generation.

\item \textit{Image-to-Panorama:} The task's target is to restore the full panorama based on the text prompt and the masked image.
Accordingly, each image is partially masked with a randomly generated mask, and GPT-5~\cite{ChatGPT} generates a full-scene description of the original (unmasked) image.
\end{itemize}

\subsubsection{Construct Input Data for Panorama Editing} To evaluate panorama editing, each test case consists of a $(\text{source image}, \text{editing prompt})$ pair.
We use the collected panorama image as the source image.
For each panorama, we define five editing tasks: \emph{removal}, \emph{addition}, \emph{replacement}, \emph{movement}, and \emph{modification}.
We employ GPT-5 to automatically detect objects, and then, for each image, generate task-specific editing instructions conditioned on the identified object categories, yielding 1{,}000 test pairs (200 images $\times$ 5 editing tasks).

\subsection{Statistical Analysis of \dataset}
\label{sec:statistical_analysis}

\begin{figure*}[t]
\centering
\includegraphics[width=1.0\linewidth]{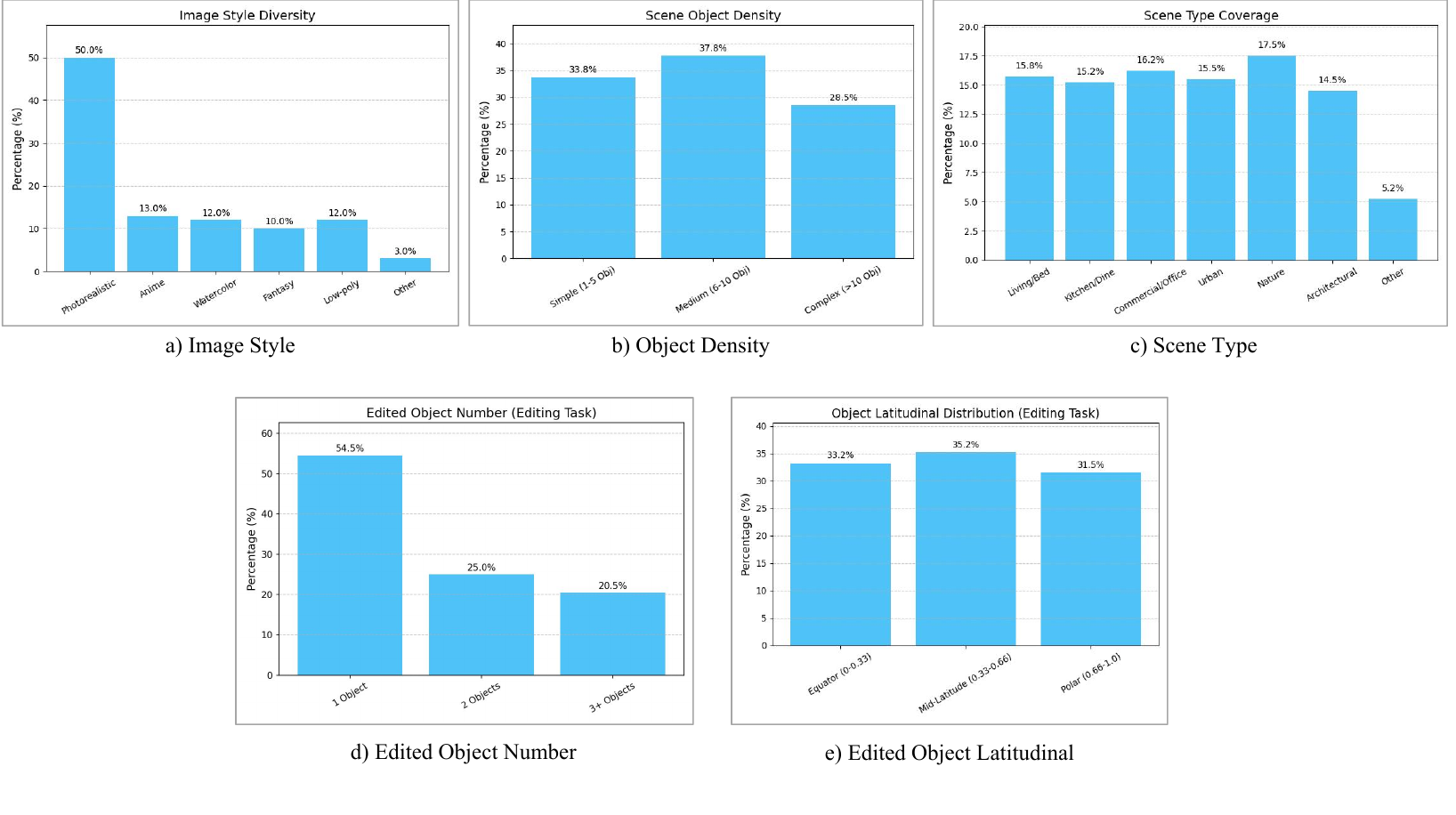}
\caption{Statistical analysis of our \dataset~ across five dimensions.}
\label{fig:bench_sta}
\end{figure*}

Our \dataset\ comprises 1,200 test samples spanning both panorama editing and generation tasks, evenly split between real and synthetic sources and between indoor and outdoor scenes. To illustrate its diversity, we provide detailed statistics across five dimensions in \figref{fig:bench_sta}(a–f).

\begin{itemize}
    \vspace{-2mm}
    \item \textbf{Image Style Diversity (\figref{fig:bench_sta}a):} The benchmark spans six major style categories. Photorealistic images constitute 50\% of the dataset, while the remaining 50\% covers non-photorealistic domains, including Anime (13\%), Watercolor (12\%), Low-poly (12\%), Fantasy (10\%), and Other styles (3\%). This balanced composition enables reliable assessment of model generalization and robustness under substantial style and domain shifts.
    \vspace{-2mm}
    \item \textbf{Scene Complexity (\figref{fig:bench_sta}b):} Scene complexity is measured based on the density of detectable objects within the panorama. Our benchmark contains scenes of varying complexity: Simple (1--5 objects, 33.8\%), \textbf{Medium} (6--10 objects, 37.8\%), and \textbf{Complex} ($>$ 10 objects, 28.5\%). This distribution supports robust evaluation across a wide range of scene complexities.
    \vspace{-2mm}
    \item \textbf{Scene Type Coverage (\figref{fig:bench_sta}c):} 
    The benchmark spans seven panorama scene types. Indoor scenes include Living/Bedroom (15.8\%), Commercial/Office (16.2\%), and Kitchen/Dining (15.2\%), while outdoor scenes cover Nature (17.5\%), Urban (15.5\%), and Architectural (14.5\%). Together with an ‘Other’ category (5.2\%), the dataset maintains a precise 50/50 indoor–outdoor split, supporting reliable evaluation across diverse interior and exterior environments.
    \vspace{-2mm}
    \item \textbf{Complexity of Editing Instructions (\figref{fig:bench_sta}d):} We assess the complexity of editing instructions in our benchmark by grouping the editing subset according to the number of targeted objects. Single-object edits represent 54.5\%, while multi-object edits constitute a substantial portion (2 objects: 25.0\%; 3+ objects: 20.5\%). This range supports thorough evaluation across editing instructions of varying complexity.
    \vspace{-2mm}
    \item \textbf{Degrees of Geometric Distortion (\figref{fig:bench_sta}e):} We assess geometric distortion in panorama editing by grouping the editing subset according to the latitudinal position of targeted objects. Edits at the equator, where distortion is minimal, account for 33.2\%, while the remaining tasks fall in regions with greater geometric distortion (mid-latitude: 35.2\%; polar: 31.5\%). This balanced distribution enables robust evaluation under varying degrees of geometric distortion.
\end{itemize}

\subsection{Training Data Construction}
\label{subsec:train_data}

To support our progressive three-stage training curriculum, we construct dedicated training datasets for each stage.

\subsubsection{Data Sources.}
Our panorama corpus is assembled from public datasets—Structured3D~\cite{zheng2020structured3d}, Pano360~\cite{SPEC:ICCV:2021}, and SUN360~\cite{xiao2012recognizing}—and augmented with custom scenes rendered in Unreal Engine (UE). These sources include both real-world captures and physically based renderings, providing panoramas with faithful equirectangular distortions. We further annotate and process this corpus to produce stage-specific training data.

\subsubsection{Training Data for Global Generation (Stage 1).}
We prepare data for two tasks focused on global structure and distortion learning:
\begin{itemize}
    \vspace{-2mm}
    \item \textit{Text-to-Panorama (T2P):} GPT-5 is used to generate descriptive text prompts for each panorama. The text serves as the input, and the original panorama is used as the ground-truth target.
    \vspace{-2mm}
    \item \textit{Image-to-Panorama (I2P):} We generate perspective images and corresponding masks by randomly projecting each panorama into perspective views. GPT-5 then produces text prompts describing the visible (unmasked) content. Each training sample consists of the perspective image, mask, and text prompt as inputs, with the full panorama as the supervision target.
\end{itemize}

\subsubsection{Training Data for Localized Controllable Generation (Stage 2).}

Stage~2 focuses on localized controllable generation. Each training sample is represented as a triplet consisting of the source panorama $I_{\text{src}}$, the control signals $\mathcal{C}_{\text{gen}}=\{P_{\text{gen}}, B, I_{\text{ref}}\}$, and the supervision targets $(I_{\text{tgt}}, L, M)$. The collected raw panoramas serve as ground-truth targets $I_{\text{tgt}}$, while all remaining components are derived through a mask-driven pipeline.

\begin{itemize}
    \vspace{-2mm}
    \item \textit{Object Shape Mask Acquisition:} We extract object masks $M$ from each panorama. For indoor scenes from Structured3D~\cite{zheng2020structured3d}, we use the provided semantic labels. For UE-rendered scenes, masks are produced automatically during rendering. For outdoor datasets (Pano360~\cite{SPEC:ICCV:2021} and SUN360~\cite{xiao2012recognizing}), we apply SAM~2~\cite{sam2} followed by manual refinement.
    
    \vspace{-2mm}
    \item \textit{Component Synthesis:} Given $M$, we generate all training components:
    (1) The \textbf{Layer} $L$ is obtained by cropping the masked region of $I_{\text{tgt}}$. (2) The \textbf{Bounding Box} $B$ is computed from the spatial extent of $M$. (3) The \textbf{Reference Image} $I_{\text{ref}}$ is created by projecting the masked region into a perspective view. (4) The \textbf{Text Prompt} $P_{\text{gen}}$ is produced using GPT-5 to describe the global scene context. (5) The \textbf{Source Panorama} $I_{\text{src}}$ serves as the background. For Structure3D~\cite{zheng2020structured3d}, we leverage the provided empty-room panoramas. For UE-rendered scenes, we obtain the background by simply removing the target object from the 3D scene and re-rendering the panorama. For the remaining real-world datasets, we synthesize the background by inpainting the perspective crop defined by $B$ and back-projecting the filled content into the panorama space.

\end{itemize}

\subsubsection{Training Data for Panorama Editing (Stage 3).}

As described in Sec.~\ref{sec:pair_data}, we construct the paired dataset $D=\{(I_{\text{src}}, P_{\text{edit}}, I_{\text{tgt}})\}$ by using the controllable generation model $G$ to synthesize $I_{\text{tgt}}$ for various editing operations.  
%
In addition to this generative pipeline, we employ a rendering-based approach using Unreal Engine (UE), where edits are applied directly to 3D assets. The modified scenes are re-rendered to produce $I_{\text{tgt}}$ with exact geometric consistency.
For all collected $(I_{\text{src}}, I_{\text{tgt}})$ pairs from both pipelines, GPT-5 compares the visual differences and automatically generates the corresponding natural-language instruction $P_{\text{edit}}$.

\section{Model Details}
\label{supp:model}

\subsection{Details on Progressive Task Transition.} To mitigate the risk of catastrophic forgetting and training instability caused by abrupt task switching, we implement a smooth curriculum transition strategy across the three training stages. Instead of hard cut-offs, we employ a dynamic data mixing protocol:
\begin{itemize}
\vspace{-3mm}
\item \textbf{Transition from Stage 1 to 2:} As the training progresses to Localized Controllable Generation (Stage 2), we do not discard the Global Structure data (Stage 1). Instead, we linearly increase the sampling probability of Stage 2 data from $0\%$ to a target ratio (e.g., $80\%$) over the initial epochs, while the remaining $20\%$ consists of Stage 1 data. This ensures that while the model learns fine-grained object control, it retains the fundamental capability of handling global spherical distortions.
\vspace{-2mm}
\item \textbf{Transition from Stage 2 to 3:} Similarly, when introducing the Supervised Editing tasks (Stage 3), we maintain containing 20\% samples from both Stage 1 (Global) and Stage 2 (Local). This strategy ensures a continuous learning trajectory, where prior capabilities serve as regularization for new tasks, ultimately yielding a unified model proficient in global generation, local control, and instruction-based editing.
\end{itemize}

\subsection{Training Details}

\noindent \textbf{Stage 1: Global Structure \& Distortion Learning.}

In implementation, we unify Text-to-Panorama (T2P) and Image-to-Panorama (I2P) by treating T2P as a special case where the mask covers the entire image ($M=\mathbf{1}$). As illustrated in Algorithm~\ref{alg:stage1}, given a target panorama $I_{\text{tgt}}$, a text prompt $P_{\text{gen}}$, and a mask $M$, we first encode the visible context $I_{\text{vis}}$ and text prompt $P_{\text{gen}}$ using Qwen2.5-VL~\cite{bai2025qwen2}, and encode the target panorama $I_{\text{tgt}}$ using VAE~\cite{vae}. We then construct the network input by concatenating the noisy latent $z_t$, the masked panorama latent $z_{\text{con}}$, and the down-sampled mask $m$. Finally, the model is optimized using a standard flow-matching denoising loss.

\begin{algorithm}[h]
  \caption{Training Process of Stage 1}
  \label{alg:stage1}
  \footnotesize
  \begin{algorithmic}
    \STATE {\bfseries Input:} Target Panorama $I_{\text{tgt}}$, Text Prompt $P_{\text{gen}}$, Mask $M$
    \STATE {\bfseries Output:} Predicted noise $\epsilon_{\text{pred}}$

    \STATE \hfill \textcolor{gray}{\textit{// Feature Encoding}}
    \STATE $I_{\text{vis}} \leftarrow I_{\text{tgt}} \odot (1 - M)$
    \STATE $c \leftarrow \mathcal{E}_{\text{Qwen}}(P_{\text{gen}}, I_{\text{vis}})$
    \STATE $z_0 \leftarrow \mathcal{E}_{\text{VAE}}(I_{\text{tgt}})$
    \STATE $m \leftarrow \text{Downsample}(M)$

    \STATE \hfill \textcolor{gray}{\textit{// Noise Injection}}
    \STATE Sample $t \sim \mathcal{U}(1, T)$ and $\epsilon \sim \mathcal{N}(0, I)$
    \STATE $z_t \leftarrow \text{Scheduler}(z_0, \epsilon, t)$

    \STATE \hfill \textcolor{gray}{\textit{// Network Input Construction}}
    \STATE $z_{\text{con}} \leftarrow z_0 \odot (1 - m)$
    \STATE $z_{\text{in}} \leftarrow \text{Concat}(z_t, z_{\text{con}}, m)$

    \STATE \hfill \textcolor{gray}{\textit{// Noise Prediction}}
    \STATE $\epsilon_{\text{pred}} \leftarrow \epsilon_\theta(z_{\text{in}}, t, c)$
  \end{algorithmic}
\end{algorithm}

\noindent \textbf{Stage 2: Localized Controllable Generation.}

To enforce position-aware shape constraints and explicitly supervise object geometry, we restructure the network output from a single panorama to a composite set containing the full target panorama $I_{\text{tgt}}$ and independent layers $\{L_{\text{tgt}}^{k}\}$ for each object.

Given the inputs, we employ Qwen2.5-VL\cite{bai2025qwen2} and VAE\cite{vae} to extract text and image features, respectively. We then inject noise into the global and layer targets, constructing the network inputs by concatenating the noisy latents with their corresponding context features, as defined in Algorithm~\ref{alg:stage2}.

Considering that processing such multi-layered inputs via simple concatenation hinders effective training due to spatial ambiguity, we follow ART~\cite{pu2025art} and employ a 3D Rotary Positional Embedding (3D-RoPE) strategy. Distinct from standard 2D RoPE, 3D-RoPE introduces an additional \texttt{layer\_id} axis alongside spatial coordinates $(x, y)$. This design enables the self-attention mechanism to efficiently distinguish between tokens belonging to the global canvas ($\texttt{layer\_id}=0$) and specific object layers ($\texttt{layer\_id}=k$).

Finally, based on the unified token sequence and global context $c$, the model simultaneously predicts the noise $v_{\text{pred}}^{\text{tgt}}$ for the global panorama and $\{v_{\text{pred}}^{(k)}\}$ for each object layer.

\begin{algorithm}[tb]
\caption{Training Process of Stage 2}
\label{alg:stage2}
\footnotesize
\begin{algorithmic}

\STATE \textbf{Input:}
Target Panorama $I_{\text{tgt}}$, Source Panorama $I_{\text{src}}$,
Layer Targets $\{L_{\text{tgt}}^{k}\}$,
Reference Images $\{I_{\text{ref}}^{k}\}$,
Box Masks $\{M_{box}^{k}\}$,
Text Prompt $P_{\text{gen}}$

\STATE \textbf{Output:}
Predicted Noise $v_{\text{pred}}^{\text{tgt}}, \{v_{\text{pred}}^{(k)}\}$

\vspace{0.8em}

\STATE \hfill \textcolor{gray}{\textit{// Feature Encoding}}
\STATE $M_{\text{union}} \leftarrow \bigcup M^{k}_{box}$
\STATE $c \leftarrow \mathcal{E}_{\text{Qwen}}(P_{\text{gen}}, I_{\text{src}}, M_{\text{union}}, \{I_{\text{ref}}^{k}\})$
\STATE $z_{\text{tgt}}, z_{\text{src}}, \{z^{k}_{layer}\}, \{z^{k}_{ref}\}
       \leftarrow \mathcal{E}_{\text{VAE}}(I_{\text{tgt}}, I_{\text{src}}, \{L_{\text{tgt}}^{k}\}, \{I_{\text{ref}}^{k}\})$
\STATE $m_{\text{union}}, \{m^{k}_{box}\} \leftarrow \text{Downsample}(M_{\text{union}}, \{M^{k}_{box}\})$

\vspace{0.8em}
\STATE \hfill \textcolor{gray}{\textit{// Noise Injection and Input Construction}}
\STATE Sample $t, \epsilon_{\text{tgt}}, \{\epsilon^{k}_{layer}\}$
\STATE $z_{t}^{\text{tgt}} \leftarrow \text{Scheduler}(z_{\text{tgt}}, \epsilon_{\text{tgt}}, t)$
\STATE $\{z_{t}^{k}\} \leftarrow \text{Scheduler}(\{z^{k}_{layer}\}, \{\epsilon^{k}_{layer}\}, t)$
\STATE $z_{\text{in}}^{0} \leftarrow \text{Concat}(z_{t}^{\text{tgt}}, z_{\text{src}} \odot (1 - m_{\text{union}}), m_{\text{union}})$
\STATE $\{z_{\text{in}}^{k}\} \leftarrow \text{Concat}(\{z_{t}^{k}\}, \{z_{\text{ref}}^{k}\}, \{m^{k}_{box}\})$

\vspace{0.8em}
\STATE \hfill \textcolor{gray}{\textit{// 3D-RoPE ID Assignment}}
\STATE $S_{0} \leftarrow \text{PatchEmbed}(z_{\text{in}}^{0})$ \textbf{with} layer\_id $= 0$
\FOR{each layer $k$}
  \STATE $S_{k} \leftarrow \text{PatchEmbed}(z_{\text{in}}^{k})$ \textbf{with} layer\_id $= k$
\ENDFOR
\STATE $S_{\text{total}} \leftarrow \text{Concat}(S_{0}, S_{1}, \dots, S_{N})$

\vspace{0.8em}
\STATE \hfill \textcolor{gray}{\textit{// Noise Prediction}}
\STATE $v_{\text{pred}}^{\text{tgt}}, \{v_{\text{pred}}^{(k)}\}
       \leftarrow \epsilon_{\theta}(S_{\text{total}}, t, c)$

\end{algorithmic}
\end{algorithm}

\noindent \textbf{Stage 3: Supervised Editing Learning.}

In the final stage, we convert the generative model into an editing model while preserving the architecture from Stage~2 to maintain consistency and retain previously learned capabilities.

We adapt the editing data to the Stage~2 input format as follows. The source image $I_{\text{src}}$ is used as the global context, and the editing instruction $P_{\text{edit}}$ serves as the text condition. Because typical editing datasets do not provide explicit object masks or reference images, these signals are replaced with null conditions, implemented analogously to Classifier-Free Guidance: the corresponding tokens or tensors are set to a learnable null embedding or zero tensor. The model’s global output target is the edited image $I_{\text{tgt}}$. For the object layers $\{L_k\}$, no ground-truth layers exist in editing pairs, so all target layers are set to transparent.

\subsection{Implementation Details}

\subsubsection{High-Resolution Adaptation}
\label{subsec:high}

To extend our framework's capability to high-resolution generation (\eg, $1024 \times 2048$), we introduce a lightweight adaptation phase. During this phase, we disable the auxiliary object layer supervision branches to reduce memory overhead and focus optimization solely on the global view. The model is then fine-tuned on high-resolution panorama images for 1,500 iterations. This strategy effectively preserves the accurate distortion priors learned in earlier stages while refining high-frequency textural details, ensuring the model produces sharp, artifact-free panoramas at higher spatial dimensions.

\subsubsection{Explicit Distortion Encoding}

To explicitly encode the distortion of ERP panoramas (with height $H$ and width $W$), which becomes increasingly severe near the poles, we introduce two complementary geometric priors.
We first introduce a \textit{Distortion Map} as extra input, $M_D \in \mathbb{R}^{H \times W}$, which encodes latitude-dependent scaling factor and tells the model the intensity of this distortion at each latitude:

\vspace{-1.5em}
\begin{equation}
M_D(x, y) = 1 - D(y) = 1 - \cos\left(\frac{\pi}{2} - \frac{\pi y}{H}\right).
\end{equation}
\vspace{-1.5em}

Second, following \cite{xia2025panowan}, we replace the standard i.i.d. Gaussian noise $\mathcal{E} \sim \mathcal{N}(0, \mathbf{I})$ with a \textit{Spatially-Distorted Noise}, $\mathcal{E}_{\text{distorted}}$. To define this noise, we first establish the latitude $\phi(y) = \frac{\pi}{2} - \frac{\pi y}{H}$ for a vertical coordinate $y \in [0, H]$, and the corresponding latitude-dependent scaling factor $D(y) = \cos(\phi(y))$. Based on the principles of spherical projection, the horizontal axis is increasingly stretched as latitude moves from the equator to the poles. 

To align our noise distribution with this geometry, we apply this same stretching principle to the noise field. We remap the horizontal sampling coordinate $x$ based on the latitude-dependent scaling factor $D(y)$:

\vspace{-1.3em}
\begin{equation}
x'(x, y) = \frac{W}{2} + \left(x - \frac{W}{2}\right) \cdot D(y).
\end{equation}
\vspace{-1.3em}

The final $\mathcal{E}_{\text{distorted}}$ is obtained normalized to maintain the  distribution on which the diffusion models are pre-trained.

\begin{figure}[!h]
    \centering
     \includegraphics[width=1.0\linewidth]{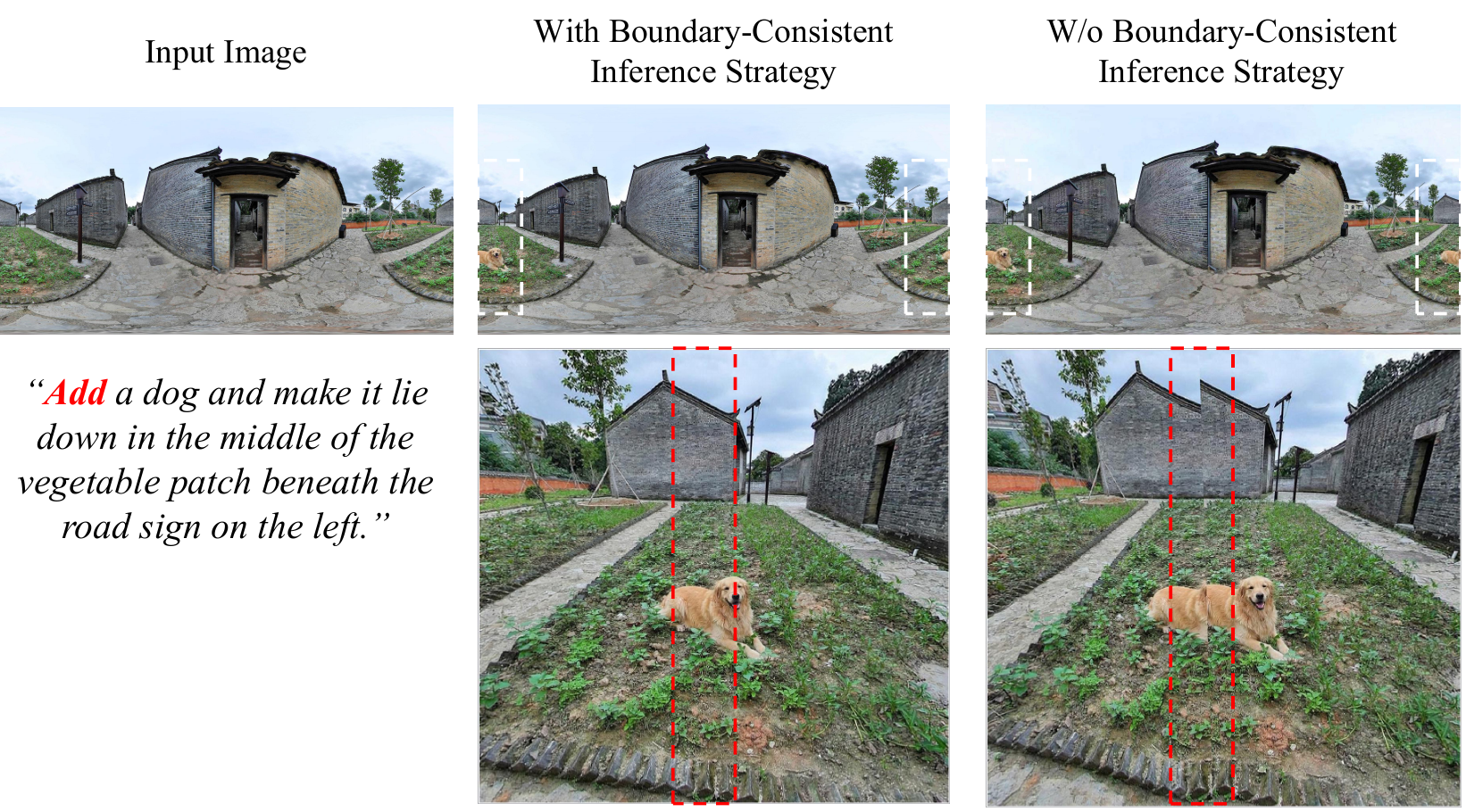}
    \caption{Using Boundary-Consistent Inference Strategy during inference time can help improve boundary continuity.}
    \label{fig:boundary}
\end{figure}

\begin{figure*}[!t]
    \centering
     \includegraphics[width=1.0\linewidth]{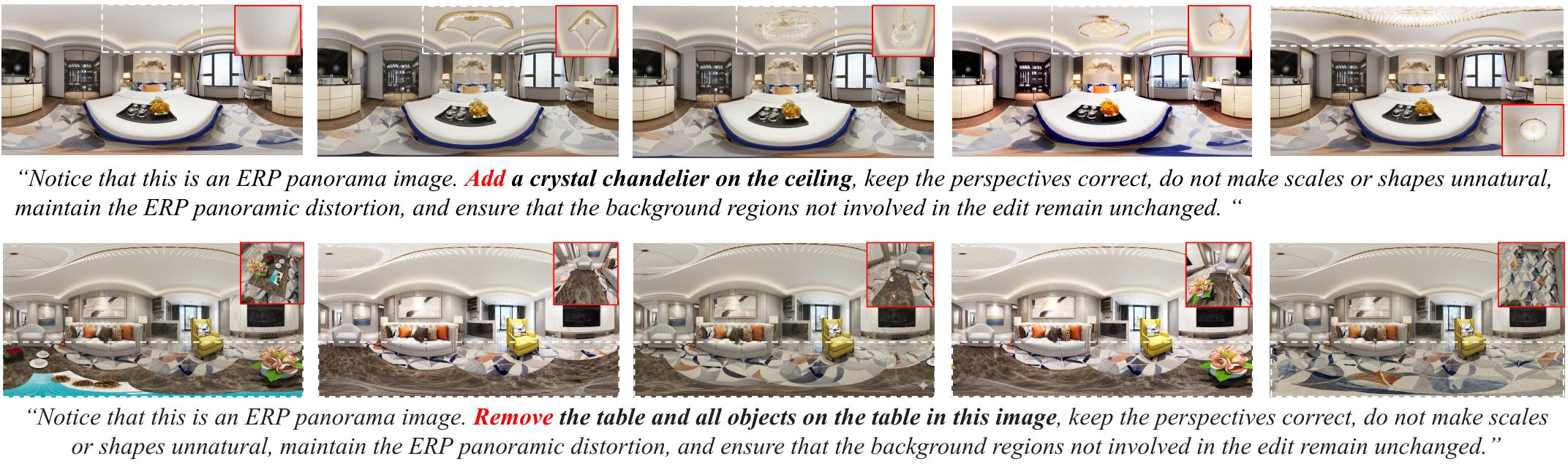}
    \put(-470,152){\small{(a) Input Image}}
    \put(-685,152){\small{\makecell{(b) Qwen\\~\cite{wu2025qwenimagetechnicalreport}}}}
    \put(-490,152){\small{\makecell{(c) Nano Banana Pro\\~\cite{Sharon2025NanoBanana}}}}
    \put(-295,152){\small{\makecell{(d) OmniGen2\\~\cite{wu2025omnigen2}}}}
    \put(-95,152){\small{\makecell{(e) Ours}}}
    \caption{Qualitative comparison with 3 top performance methods from Table.~\ref{tab:sota} using enhanced text prompt.}
    \label{fig:prompt}
\end{figure*}

\begin{table*}[!t]
\centering
\small
\caption{Quantitative comparison of our method against eight SOTA image editing methods using prompt enhancement by GPT-5~\cite{ChatGPT}.
We report automatic evaluation metrics on PEBench using enhanced prompts for all baselines.
The best and second-best results are highlighted in \textbf{bold} and \underline{underlined}, respectively.
$^{\dag}$ denotes proprietary methods which cannot be re-trained.
All CLIP$_{dir}$ scores are multiplied by 100 for readability.
Numbers indicate the performance after prompt enhancement, with relative improvements over the original prompts shown in parentheses.}
\vspace{-2mm}
\setlength{\tabcolsep}{3pt}
\renewcommand{\arraystretch}{1.25}
\scalebox{0.65}{
\begin{tabular}{l|l|ccccccccc}
\toprule
\multirow{1}{*}{\textbf{Category}} & \multirow{1}{*}{\textbf{Metrics}}
& \makecell{Qwen\\~\cite{wu2025qwenimagetechnicalreport}}
& \makecell{Kontext\\~\cite{batifol2025flux}}
& \makecell{GPT-5$^{\dag}$\\~\cite{ChatGPT}}
& \makecell{Nano Banana Pro$^{\dag}$\\~\cite{Sharon2025NanoBanana}}
& \makecell{IC-Edit\\~\cite{zhang2025context}}
& \makecell{Omni$^2$\\~\cite{omni2}}
& \makecell{Step1X-Edit\\~\cite{liu2025step1x}}
& \makecell{OmniGen2\\~\cite{wu2025omnigen2}}
& Ours \\
\midrule

\multirow{5}{*}{Automatic}

& FID $\downarrow$
& 63.32~(-2.10)
& 58.73~(-1.45)
& 88.57~(-3.80)
& \underline{50.15}~(-0.95)
& 82.08~(-2.75)
& 59.95~(-1.20)
& 75.59~(-3.10)
& 64.48~(-2.05)
& \textbf{44.65}~(-0.60) \\

& $\text{CLIP}_{\text{dir}}$ $\uparrow$
& 14.99~(+0.42)
& 13.38~(+0.33)
& 13.20~(+0.28)
& \underline{17.25}~(+0.51)
& 16.68~(+0.47)
& 17.04~(+0.39)
& 12.18~(+0.25)
& 13.55~(+0.31)
& \textbf{19.97}~(+0.35) \\

& SC $\uparrow$
& 6.09~(+0.18)
& 3.71~(+0.09)
& 2.59~(+0.06)
& 6.03~(+0.15)
& 5.88~(+0.14)
& 6.03~(+0.17)
& 5.53~(+0.12)
& \underline{6.18}~(+0.20)
& \textbf{8.27}~(+0.12) \\

& PQ $\uparrow$
& 5.58~(+0.16)
& 4.14~(+0.11)
& 5.04~(+0.13)
& 5.56~(+0.15)
& 5.37~(+0.14)
& 5.48~(+0.12)
& \underline{5.66}~(+0.18)
& 5.21~(+0.10)
& \textbf{8.02}~(+0.15) \\

& O $\uparrow$
& \underline{5.83}~(+0.17)
& 2.85~(+0.08)
& 2.80~(+0.07)
& 5.80~(+0.16)
& 5.63~(+0.15)
& 5.74~(+0.14)
& 5.57~(+0.13)
& 5.65~(+0.12)
& \textbf{8.12}~(+0.11) \\

\bottomrule
\end{tabular}}
\label{tab:sota_enhance_prompt}
\vspace{0mm}
\end{table*}

\subsubsection{Boundary-Consistent Inference Strategy}

Our method performs panorama editing directly in the equirectangular projection (ERP) domain. 
An important property of ERP panoramas is that they are intrinsically periodic and continuous along the longitudinal (horizontal) direction, 
where the left and right image boundaries correspond to adjacent regions on the sphere.

\begin{algorithm}[h]
  \caption{Boundary-Consistent Inference Strategy}
  \label{alg:boundary_editing}
  \footnotesize
  \begin{algorithmic}
    \STATE {\bfseries Input:} Source Panorama $I_{src}$, Edit Instruction $P_{\text{edit}}$, Model $\epsilon_\theta$, Scheduler $\mathcal{S}$, VAE Encoder/Decoder $\mathcal{E}_{\text{VAE}}, \mathcal{D}_{\text{VAE}}$
    \STATE {\bfseries Hyper-params:} Boundary extension width $b$, Shifting offset $s$, Shift steps $K$, Denoise steps $T$
    \STATE {\bfseries Output:} Boundary-consistent Edited Panorama $\hat{I_{tgt}}$

    \STATE \hfill \textcolor{gray}{\textit{// Encode panorama into latent space}}
    \STATE $z_0 \leftarrow \mathcal{E}_{\text{VAE}}(I)$

    \STATE \hfill \textcolor{gray}{\textit{// Circular boundary extension}}
    \STATE $\tilde{z}_0 \leftarrow \text{Concat}\big(z_0,\; z_0[:, :, :, 1{:}b]\big)$

    \STATE \hfill \textcolor{gray}{\textit{// Initialize noisy latent}}
    \STATE Sample $\epsilon \sim \mathcal{N}(0, I)$
    \STATE $\tilde{z}_T \leftarrow \mathcal{S}(\tilde{z}_0, \epsilon, T)$

    \STATE \hfill \textcolor{gray}{\textit{// Denoising loop with early-stage shifting}}
    \FOR{$t = T, T\!-\!1, \dots, 1$}
        \STATE $\epsilon_{\text{pred}} \leftarrow \epsilon_\theta(\tilde{z}_t, t, P_{\text{edit}})$
        \STATE $\tilde{z}_{t-1} \leftarrow \mathcal{S}(\tilde{z}_t, \epsilon_{\text{pred}}, t)$

        \IF{$t > T-K$}
            \STATE \hfill \textcolor{gray}{\textit{// Cyclic shifting for seam suppression}}
            \STATE $\tilde{z}_{t-1} \leftarrow \text{Roll}(\tilde{z}_{t-1}, s, \text{horiz})$
        \ENDIF
    \ENDFOR

    \STATE \hfill \textcolor{gray}{\textit{// Boundary blending + crop back}}
    \STATE $\tilde{z}_0^\star \leftarrow \text{BlendBoundary}(\tilde{z}_0, b)$
    \STATE $z_0^\star \leftarrow \tilde{z}_0^\star[:, :, :, 1{:}W]$

    \STATE \hfill \textcolor{gray}{\textit{// Decode edited panorama}}
    \STATE $\hat{I} \leftarrow \mathcal{D}_{\text{VAE}}(z_0^\star)$
  \end{algorithmic}
\end{algorithm}

To improve the continuity, we use a \emph{Boundary-Consistent Inference Strategy} that explicitly enforces an approximate periodic boundary condition. 
As illustrated in Alg.~\ref{alg:boundary_editing}, we first perform \emph{Circular boundary extension} by concatenating a narrow strip from the left boundary to the right side in the latent width dimension. 
This enables the model to access the true neighboring context across the wrap-around boundary during denoising. 
Next, during the early diffusion steps, we apply a cyclic horizontal shift (\textit{Roll}) to the latent features, 
which effectively shifts the seam location at the beginning of generation and prevents artifacts from consistently accumulating along a fixed vertical boundary. 
Finally, we apply \emph{Boundary blending} over the extended band and crop the latent back to the original width, further suppressing pixel-domain boundary artifacts and improving left-right continuity.

\section{Additional Experiments \& Results}
\label{supp:exp}

\subsection{Prompt Enhancement}
\label{sec:prompt_enhance}

In Section~\ref{sec:qual_comp}, we observe that current baseline methods suffer from two fundamental limitations when applied to ERP panoramic image editing: 
\textbf{Distortion Unawareness} and \textbf{Background Instability}. 
A possible explanation is that these failures may be caused by suboptimal text prompts, since such models heavily rely on language instructions.

To rule out this factor, we further strengthen the prompts by explicitly providing additional cues, encouraging the models to recognize that the input is an \textit{ERP panorama image}, to follow the corresponding geometric distortions, and to preserve the background regions that are not involved in the edit.

For example, for the \textit{addition} task, we enhance the prompt as follows: \textit{"Notice that this is an ERP panorama image. Add X to Y, keep the perspectives correct, do not make scales or shapes unnatural, maintain the ERP panoramic distortion, and ensure that the background regions not involved in the edit remain unchanged."}

Fig.~\ref{fig:prompt} shows that even with these prompt enhancements, existing baseline methods still fail to achieve correct edits. Furthermore, we also adopt an LLM (e.g., GPT) to automatically enhance the prompts by injecting additional cues, such as explicitly indicating that the input is a panoramic image, that the ERP distortion should be preserved, and that background regions unrelated to the editing target should remain unchanged. We then re-evaluate these baselines on \dataset.

As shown in Table~\ref{tab:sota_enhance_prompt}, the reported results are obtained using enhanced prompts, with the numbers in parentheses indicating the relative improvement over standard prompts. While prompt enhancement can moderately improve the performance of baseline models, generating geometrically correct distortions remains challenging for these methods.

\begin{figure*}[t]
    \centering
     \includegraphics[width=1.0\linewidth]{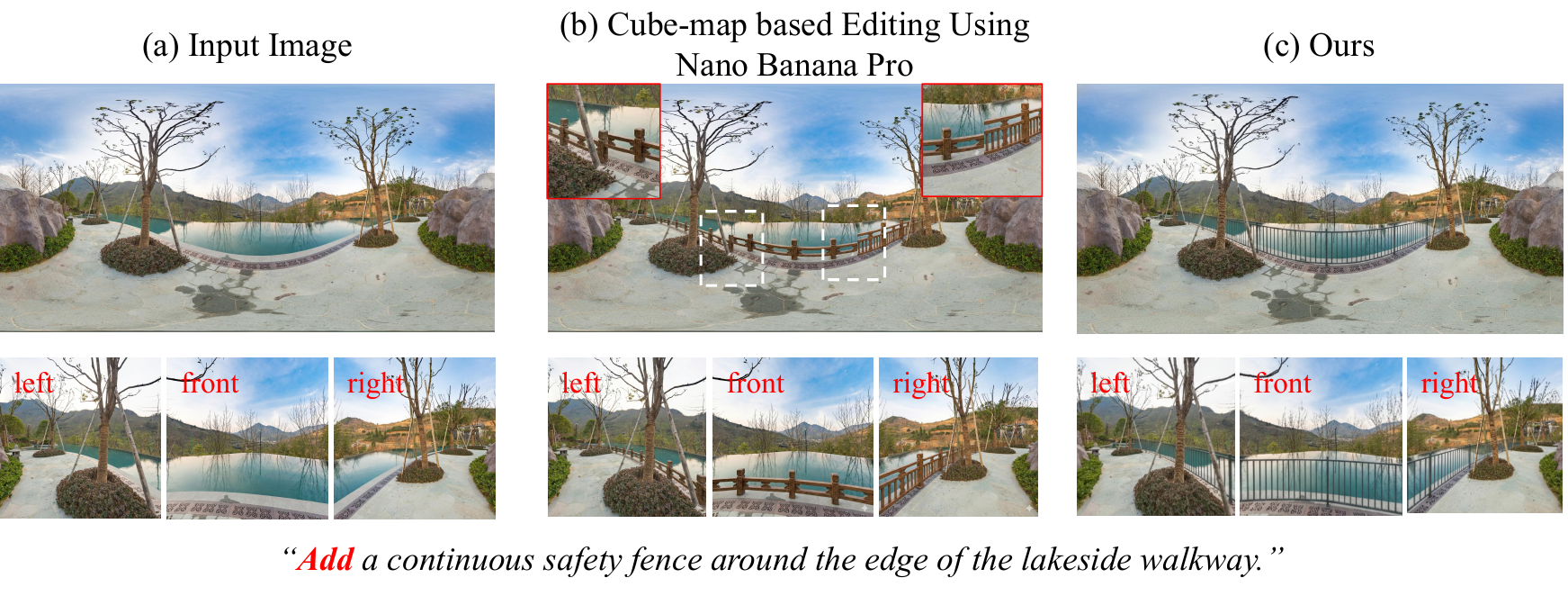}
    \caption{When editing objects spanning multiple cube faces, cube-map based pipelines may introduce noticeable artifacts at face boundaries.}
    \label{fig:cubevserp}
\end{figure*}

\subsection{Advantages of Using ERP-native Representation}
\label{sec:more_motivation}

One simple workaround for panorama editing is to first locate the target object within a specific cube face, apply a perspective image editing model on that cube, and then project the edited result back to the ERP panorama, this cube-based pipeline suffers from several inherent limitations.

\textbf{Difficulty in editing cross-face objects.} First, panoramic scenes often contain objects that span a wide field of view. As illustrated in Fig.~\ref{fig:cubevserp} (b), when the target object extends across multiple cube faces, it becomes difficult to generate satisfactory results. In such cases, the object must be edited separately on several cube views (e.g., left, front, and right). This fragments the object into multiple parts, breaking its geometric continuity. Consequently, noticeable artifacts frequently appear at cube boundaries when the edited faces are stitched back together (Fig.~\ref{fig:cubevserp} (b) ). In contrast, our ERP-based editing framework operates directly on the continuous panoramic representation, avoiding such segmentation artifacts.

\textbf{Challenges in global style editing.} Second, cube-based approaches are not well suited for global style editing. Since the panorama is processed as disjoint perspective views, enforcing consistent appearance changes across the entire scene becomes challenging. Our ERP-based method naturally supports global style edits over the full panoramic image, as demonstrated in Fig.~\ref{fig:global_style}.

\textbf{Complicated multi-stage pipeline.} Third, cube-based editing pipelines involve a complicated multi-stage procedure. 
They require decomposing the panorama into cube faces, selecting the relevant views, performing edits on each face separately, and finally projecting and stitching the results back to the ERP domain. 
This process becomes even more cumbersome for multi-object editing, as multiple faces may need to be edited sequentially. 
In contrast, our ERP-based method supports editing multiple objects directly in a single forward pass, avoiding repeated decomposition and reprojection steps.

To further validate the advantage of directly operating on ERP panoramas, we conduct an additional experiment by decomposing each ERP training image into six cube faces, and retraining the baseline methods using cube inputs instead of ERP images. 

\begin{table}[h]
\centering
\footnotesize
\caption{Comparison between ERP and cube-map inputs for representative editing baselines on \textbf{panorama editing}. The best results are highlighted in \textbf{bold}.}

\setlength{\tabcolsep}{6pt}
\renewcommand{\arraystretch}{1.25}
\scalebox{0.65}{
\begin{tabular}{l|ccccc}
\toprule
\textbf{Method} & FID $\downarrow$ & CLIP$_{dir}$ $\uparrow$ & SC $\uparrow$ & PQ $\uparrow$ & O $\uparrow$ \\
\midrule

Qwen\cite{wu2025qwenimage} (ERP Input)      & 65.42 & 14.57 & 5.91 & 5.42 & 5.66 \\
Qwen\cite{wu2025qwenimage} (Cube-map Input) & 73.92 & 12.82 & 5.14 & 4.66 & 4.98 \\

\midrule
Kontext\cite{batifol2025flux1kontext} (ERP Input)      & 60.18 & 13.05 & 3.62 & 4.03 & 2.77 \\
Kontext\cite{batifol2025flux1kontext} (Cube-map Input) & 68.61 & 11.22 & 3.22 & 3.51 & 2.35 \\

\midrule
OmniGen2\cite{wu2025omnigen2} (ERP Input)      & 66.53 & 13.24 & 5.98 & 5.11 & 5.53 \\
OmniGen2\cite{wu2025omnigen2} (Cube-map Input) & 75.18 & 11.52 & 5.14 & 4.50 & 4.76 \\

\midrule
Ours (ERP Input) & \textbf{45.25} & \textbf{19.62} & \textbf{8.15} & \textbf{7.87} & \textbf{8.01} \\

\bottomrule
\end{tabular}}
\label{tab:erp_vs_cube}
\vspace{-5mm}
\end{table}

As shown in Table.~\ref{tab:erp_vs_cube}, models trained with ERP inputs consistently achieve better performance than those trained with cubemap inputs across all evaluation metrics. This suggests that ERP inputs may better align with the priors of pretrained editing models. In contrast, cubemap-based training could introduce asymmetric face-wise distortions and discontinuities, which may disrupt spatial consistency and increase optimization difficulty.

\begin{table*}[t!]
\centering
\small
\caption{Quantitative comparison against SOTA methods on \textbf{Text-to-Panorama} generation. The user study reports the average ranking in image quality (IQ), distortion accuracy (DA), and text alignment (TA).  
The best and second-best results are highlighted in \textbf{bold} and \underline{underlined}, respectively. 
}
\vspace{-2mm}
\setlength{\tabcolsep}{2pt}
\renewcommand{\arraystretch}{1.1}
\scalebox{0.9}{
\begin{tabular}{l|l|ccccccc}
\toprule
\textbf{Category} 
& \textbf{Metrics}
& \makecell{UniPano\\{\cite{ni2025makes}}}
& \makecell{Panfusion\\{\cite{zhang2024taming}}}
& \makecell{SMGD\\{\cite{sun2025spherical}}}
& \makecell{Diff360\\{\cite{feng2023diffusion360}}}
& \makecell{MVDiff\\{\cite{tang2023mvdiffusionenablingholisticmultiview}}}
& \makecell{DiT360\\{\cite{feng2025dit360}}}
& \makecell{\textbf{Ours}} \\
\midrule
\multirow{4}{*}{Automatic}
 & FID $\downarrow$        
 & 57.53 & 81.01 & 69.25 & 88.09 & 96.07 & \underline{51.42} & \textbf{48.39} \\
 & CLIP-Score $\uparrow$  
 & 28.03 & 28.10 & \underline{28.81} & 27.59 & 27.71 & 28.45 & \textbf{28.89} \\
 & IS $\uparrow$           
 & 4.02 & 3.91 & 3.95 & 3.35 & 3.13 & \underline{4.15} & \textbf{4.45} \\
 & FAED $\downarrow$       
 & \underline{8.15} & 9.44 & 9.02 & 11.22 & 15.35 & 8.35 & \textbf{7.21} \\
\midrule
\multirow{3}{*}{User study}
 & IQ $\downarrow$            
 & 3.57 & 5.14 & 4.88 & 5.21 & 5.65 & \underline{2.42} & \textbf{1.13} \\
 & DA $\downarrow$            
 & \underline{2.62} & 4.73 & 5.68 & 5.09 & 5.74 & 2.98 & \textbf{1.16} \\
 & TA $\downarrow$            
 & 4.48 & 3.12 & \underline{2.28} & 6.18 & 6.58 & 4.12 & \textbf{1.24} \\
\bottomrule
\end{tabular}}
\label{tab:sota_gene_text}
\vspace{3mm}
\end{table*}

\begin{table*}[t!]
\centering
\small
\caption{Quantitative comparison against SOTA methods on \textbf{Image-to-Panorama} generation. 
The user study reports the average ranking in image quality (IQ), distortion accuracy (DA), and text alignment (TA).  
The best and second-best results are highlighted in \textbf{bold} and \underline{underlined}, respectively. 
}
\vspace{-2mm}
\setlength{\tabcolsep}{2pt}
\renewcommand{\arraystretch}{1.1}
\scalebox{0.82}{
\begin{tabular}{l|l|ccccccc}
\toprule
\textbf{Category}
& \textbf{Metrics}
& \makecell{CubeDiff\\{\cite{kalischek2025cubediff}}}
& \makecell{DreamCube\\{\cite{huang2025dreamcube}}}
& \makecell{PanoDiff\\{\cite{wang2023360}}}
& \makecell{OmniX\\{\cite{huang2025omnix}}}
& \makecell{PanoDiffusion\\{\cite{wu2023panodiffusion}}}
& \makecell{OmniDrmr\\{\cite{akimoto2022diverse}}}
& \makecell{\textbf{Ours}} \\
\midrule
\multirow{4}{*}{Automatic}
 & FID $\downarrow$        
 & 53.76 & \underline{50.15} & 76.45 & 58.12 & 89.45 & 111.12 & \textbf{46.14} \\
 & CLIP-Score $\uparrow$  
 & 28.15 & 28.42 & 27.88 & \underline{28.62} & 27.55 & 27.58 & \textbf{28.93} \\
 & IS $\uparrow$           
 & 3.89 & \underline{4.25} & 3.65 & 4.11 & 3.32 & 2.77 & \textbf{4.71} \\
 & FAED $\downarrow$       
 & \underline{7.45} & 7.92 & 9.14 & 8.45 & 17.21 & 19.82 & \textbf{7.09} \\
\midrule
\multirow{3}{*}{User study}
 & IQ $\downarrow$            
 & 3.65 & \underline{2.35} & 5.35 & 2.55 & 6.15 & 6.80 & \textbf{1.15} \\
 & DA $\downarrow$            
 & \underline{2.22} & 2.32 & 5.25 & 3.83 & 6.35 & 6.85 & \textbf{1.18} \\
 & TA $\downarrow$            
 & 3.98 & 2.62 & 5.15 & \underline{2.28} & 6.10 & 6.63 & \textbf{1.24} \\
\bottomrule
\end{tabular}}
\label{tab:sota_gene_image}
\vspace{3mm}
\end{table*}

\begin{figure*}[t!]
    \centering
    \includegraphics[width=1.0\linewidth]{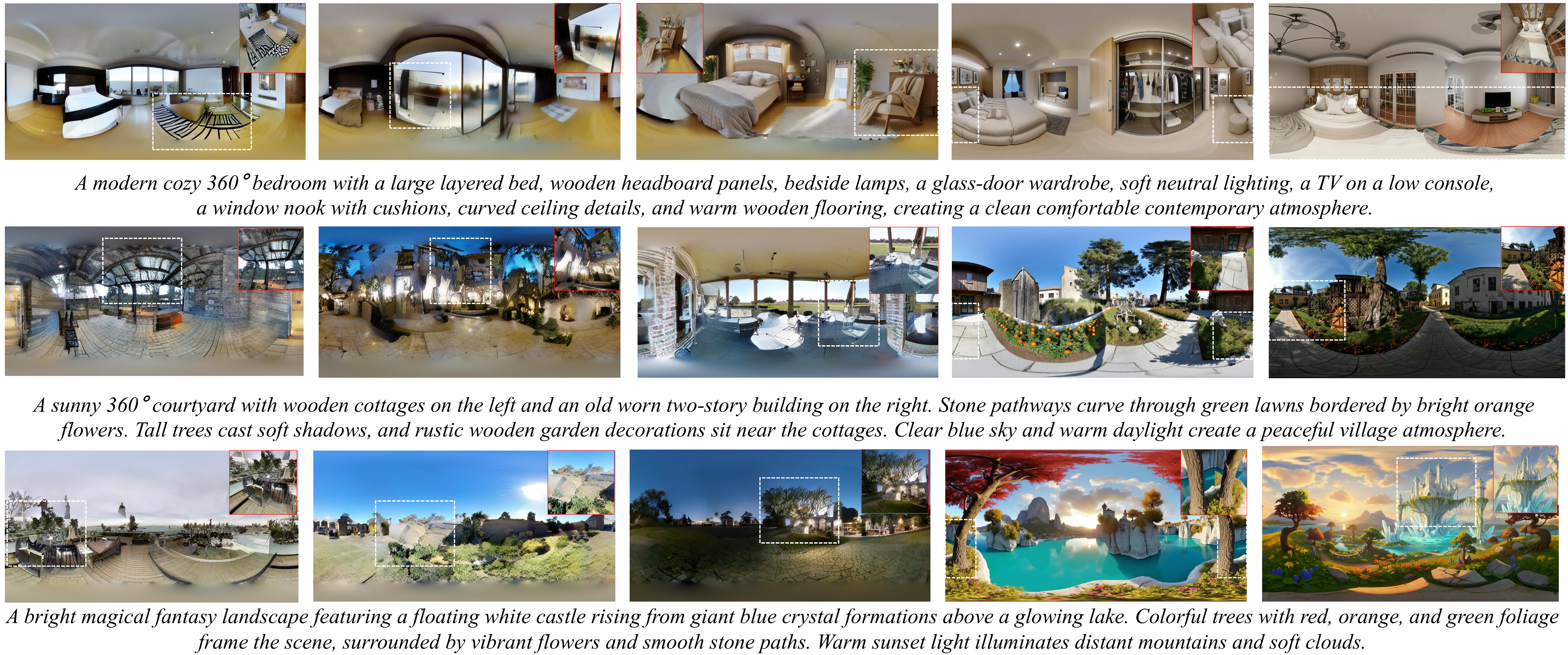}
    \put(-875,213){\makecell{\small (a) SMGD \\ \cite{sun2025spherical}}}
    \put(-680,213){\makecell{\small (b) PanFusion \\ \cite{zhang2024taming}}}
    \put(-485,213){\makecell{\small (c) UniPano \\ \cite{ni2025makes}}}
    \put(-290,213){\makecell{\small (d) DiT360 \\ \cite{feng2025dit360}}}
    \put(-100,210){\makecell{\small (e) Ours}}
    \caption{Qualitative comparison on Text-to-Panorama generation with top performance methods in Tab.~\ref{tab:sota_gene_text}.}
    \label{fig:compare_sota_t2i}
\end{figure*}

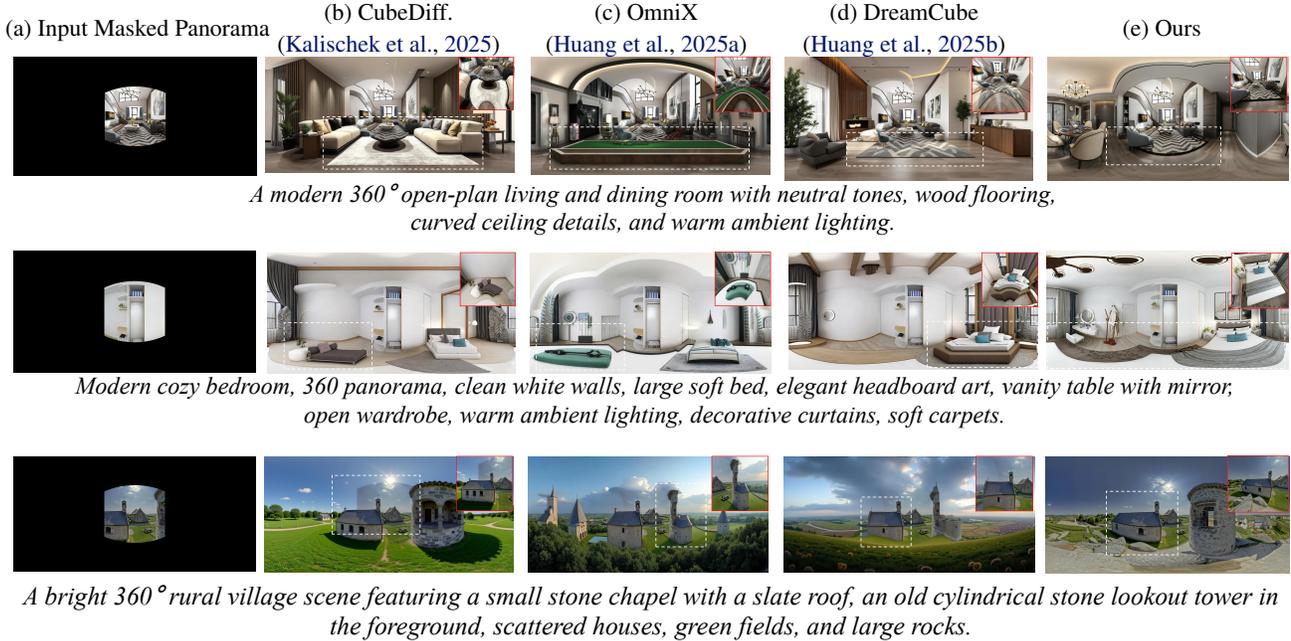
\begin{figure*}[t!]
    \centering
    \includegraphics[width=1.0\linewidth]{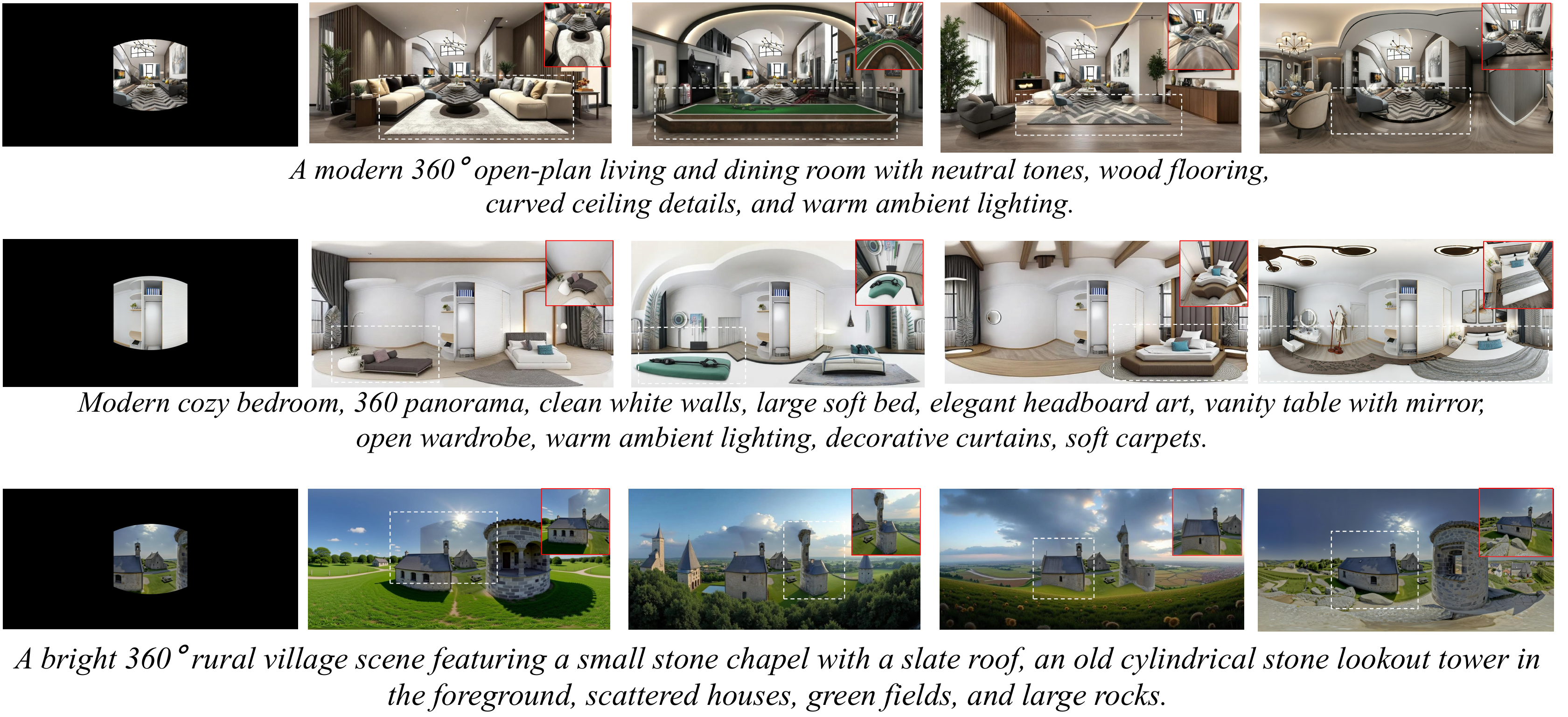}
    \put(-880,230){\makecell{\small (a) Input Masked Panorama}}
    \put(-690,230){\makecell{\small (b) CubeDiff. \\ \small \cite{kalischek2025cubediff}}}
    \put(-495,230){\makecell{\small (c) OmniX \\ \small \cite{huang2025omnix}}}
    \put(-300,230){\makecell{\small (d) DreamCube \\ \small \cite{huang2025dreamcube}}}
    \put(-105,230){\makecell{\small (e) Ours}}
    \caption{Qualitative comparison Image-to-Panorama generation with top performance methods in Tab.~\ref{tab:sota_gene_image}.}
    \label{fig:compare_sota_i2i}
\end{figure*}

\subsection{Comparisons on Panoramic Image Generation}
\label{subsubsec:comparison_editing}

We further evaluate our method on panorama generation.
For text-to-panorama generation, we compare 
\name~ with 6 SOTA methods: UniPano~\cite{ni2025makes}, 
 Panfusion~\cite{zhang2024taming}, SMGD~\cite{sun2025spherical}, Diffusion360~\cite{feng2023diffusion360}, MVDiffusion~\cite{tang2023mvdiffusionenablingholisticmultiview} and DiT360\cite{feng2025dit360}.
 
For image-to-panorama generation: we compare \name~with 6 SOTA methods: 
against PanoDiff\cite{wang2023360}, OmniX~\cite{huang2025omnix}, 
DreamCube\cite{huang2025dreamcube}, 
CubeDiff\cite{kalischek2025cubediff}, PanoDiffusion~\cite{wu2023panodiffusion}, and OmniDreamer~\cite{akimoto2022diverse}.

\noindent \textbf{Generation Metrics:} We employ Fréchet Inception Distance (FID)~\cite{heusel2017gans}, Inception Score (IS)~\cite{salimans2016improved}, and Fréchet Auto-Encoder Distance (FAED)~\cite{oh2022bips,zhang2024taming} to measure realism, diversity and geometric consistency. Again, CLIP-Score is also utilized to measure the text-image alignment. 

\noindent \textbf{Quantitative Comparisons.}
Table~\ref{tab:sota_gene_text} and Table~\ref{tab:sota_gene_image} reports quantitative results for both text-to-panorama and image-to-panorama generation. Our method consistently outperforms all baselines across automatic metrics and user evaluation.

\noindent \textbf{Qualitative Comparison.} We also provide visual comparisons in \figref{fig:compare_sota_t2i} and \figref{fig:compare_sota_i2i}. 
Existing methods either fail at text–image alignment, 
or exhibit noticeable geometric inconsistencies, including edge artifacts and distortions. Our method avoids these issues, delivering realistic, distortion-consistent, and instruction-aligned results.

\noindent{\textbf{User Study.}} We further conduct a user study with 65 participants to assess human preferences across both text-to-panorama and image-to-panorama generation. 
Each participant is shown 20 text-to-panorama cases (a text prompt with shuffled results from all competing methods) and 20 image-to-panorama cases (a masked input panorama with shuffled outputs).  
For each case, participants rank all methods along three criteria: image quality (IQ), distortion accuracy (DA), and text alignment (TA).
We aggregate rankings across all participants and report the averaged scores in the bottom rows of Table~\ref{tab:sota_gene_text} and Table~\ref{tab:sota_gene_image}. 
These results show that our method is consistently ranked highest on all three aspects.

\subsection{More Evaluation on Public Available Datasets}
\label{sec:more_public_eval}

To provide a more comprehensive and convincing evaluation besides ~\dataset, we further conduct experiments on several publicly available datasets. These additional results aim to better demonstrate the generalization ability of our method under diverse settings and larger evaluation scales.

\subsubsection{Evaluation on the Public Test Set of Concurrent Work}

First, we evaluate our method on the public test set released by concurrent work SE360~\cite{SE360}, which contains \textbf{660} test cases.
Since this dataset is constructed independently and is publicly accessible, it serves as a reliable benchmark for evaluating performance beyond our curated benchmark. As shown in Table~\ref{tab:se360_benchmark}, our method achieves the best performance on this benchmark across all evaluation metrics.

\begin{table*}[t!]
\centering
\small
\caption{Quantitative comparison on the \textbf{SE360's~\cite{SE360} Benchmark}. 
Our method is evaluated against eight SOTA image editing baselines for panorama tasks.
The best and second-best results are highlighted in \textbf{bold} and \underline{underlined}, respectively.
$^{\dag}$ denotes proprietary methods.
All CLIP$_{dir}$ scores are scaled ($\times 100$) for readability.}
\vspace{-2mm}
\setlength{\tabcolsep}{3pt}
\renewcommand{\arraystretch}{1.1}
\scalebox{0.68}{
\begin{tabular}{l|l|ccccccccc}
\toprule
\multirow{1}{*}{\textbf{Category}} & \multirow{1}{*}{\textbf{Metrics}}
& \makecell{Qwen\\~\cite{wu2025qwenimagetechnicalreport}}
& \makecell{Kontext\\~\cite{batifol2025flux}}
& \makecell{GPT-5$^{\dag}$\\~\cite{ChatGPT}}
& \makecell{Nano Banana Pro$^{\dag}$\\~\cite{Sharon2025NanoBanana}}
& \makecell{IC-Edit\\~\cite{zhang2025context}}
& \makecell{SE360\\~\cite{SE360}}
& \makecell{Step1X-Edit\\~\cite{liu2025step1x}}
& \makecell{OmniGen2\\~\cite{wu2025omnigen2}}
& Ours \\
\midrule
\multirow{5}{*}{Automatic}
& FID $\downarrow$  
  & 67.21 & 58.94 & 88.42 & \underline{49.82} & 81.36 & 62.47 & 76.15 & 64.92 & \textbf{43.18} \\
& $\text{CLIP}_{\text{dir}}$ $\uparrow$  
  & 15.02 & 13.41 & 12.18 & \underline{17.22} & 15.89 & 16.14 & 12.44 & 13.91 & \textbf{20.55} \\
& SC $\uparrow$  
  & 5.64 & 3.48 & 2.62 & 5.92 & 5.51 & 6.02 & 5.12 & \underline{6.14} & \textbf{8.43} \\
& PQ $\uparrow$  
  & 5.28 & 4.21 & 4.65 & 5.54 & 5.08 & 5.49 & \underline{5.62} & 4.98 & \textbf{8.12} \\
& O $\uparrow$  
  & \underline{5.82} & 2.91 & 2.85 & 5.51 & 5.66 & 5.34 & 5.27 & 5.71 & \textbf{8.34} \\
\bottomrule
\end{tabular}}
\label{tab:se360_benchmark}
\vspace{0mm}
\end{table*}

\begin{table*}[t!]
\centering
\small
\caption{Quantitative results on the combined \textbf{Structured3D}~\cite{zheng2020structured3d} and \textbf{SUN360}~\cite{xiao2012recognizing} datasets. 
Our method demonstrates consistent superiority in panorama editing compared to existing SOTA models.
The best and second-best results are highlighted in \textbf{bold} and \underline{underlined}, respectively.
$^{\dag}$ denotes proprietary methods.
CLIP$_{dir}$ scores are scaled by 100.}
\vspace{-2mm}
\setlength{\tabcolsep}{3pt}
\renewcommand{\arraystretch}{1.1}
\scalebox{0.68}{
\begin{tabular}{l|l|ccccccccc}
\toprule
\multirow{1}{*}{\textbf{Category}} & \multirow{1}{*}{\textbf{Metrics}}
& \makecell{Qwen\\~\cite{wu2025qwenimagetechnicalreport}}
& \makecell{Kontext\\~\cite{batifol2025flux}}
& \makecell{GPT-5$^{\dag}$\\~\cite{ChatGPT}}
& \makecell{Nano Banana Pro$^{\dag}$\\~\cite{Sharon2025NanoBanana}}
& \makecell{IC-Edit\\~\cite{zhang2025context}}
& \makecell{SE360\\~\cite{SE360}}
& \makecell{Step1X-Edit\\~\cite{liu2025step1x}}
& \makecell{OmniGen2\\~\cite{wu2025omnigen2}}
& Ours \\
\midrule
\multirow{5}{*}{Automatic}
& FID $\downarrow$  
  & 68.45 & 61.22 & 85.19 & \underline{47.36} & 79.54 & 64.10 & 74.28 & 63.81 & \textbf{41.05} \\
& $\text{CLIP}_{\text{dir}}$ $\uparrow$  
  & 15.68 & 14.12 & 13.05 & \underline{18.41} & 16.32 & 17.02 & 13.56 & 14.23 & \textbf{21.37} \\
& SC $\uparrow$  
  & 5.42 & 3.15 & 2.88 & 6.07 & 5.29 & 6.18 & 5.33 & \underline{6.45} & \textbf{8.76} \\
& PQ $\uparrow$  
  & 5.11 & 4.38 & 4.72 & 5.69 & 5.34 & 5.51 & \underline{5.84} & 5.10 & \textbf{8.35} \\
& O $\uparrow$  
  & \underline{5.94} & 3.06 & 2.92 & 5.72 & 5.81 & 5.46 & 5.50 & 5.89 & \textbf{8.62} \\
\bottomrule
\end{tabular}}
\label{tab:s3d_sun360_results}
\vspace{0mm}
\end{table*}

\subsubsection{Large-Scale Evaluation on Structured3D and SUN360}

To further increase the evaluation scale and reduce potential benchmark-specific bias, we additionally build a large-scale test set based on two widely-used public datasets, \textbf{Structured3D} (indoor scenes) and \textbf{SUN360} (outdoor panoramic scenes).

While this supplementary benchmark provides a clear advantage in terms of data volume, ~\dataset remains necessary due to its richer editing-oriented design, featuring more diverse scene content (e.g., stylized and UE-rendered panoramas), together with careful manual inspection to ensure the appropriateness and accuracy of the editing instructions.

We randomly sample \textbf{1,600} images from each public dataset’s \textbf{test split}, resulting in a total of \textbf{3,200} test images, covering both indoor and outdoor environments, which do not overlap with any training data.

For each image, we generate an editing instruction using GPT, following the same procedure described in Sec.~\ref{subsubsec:dataset} for \dataset. 
We then evaluate all methods under identical experimental settings on this large-scale benchmark. 
The results in \tableref{tab:s3d_sun360_results} show that our approach consistently outperforms strong baselines, confirming that the improvements generalize beyond small-scale curated benchmarks.

\subsection{Global Style Edit}

\begin{figure*}[!h]
    \centering
    \includegraphics[width=1.0\linewidth]{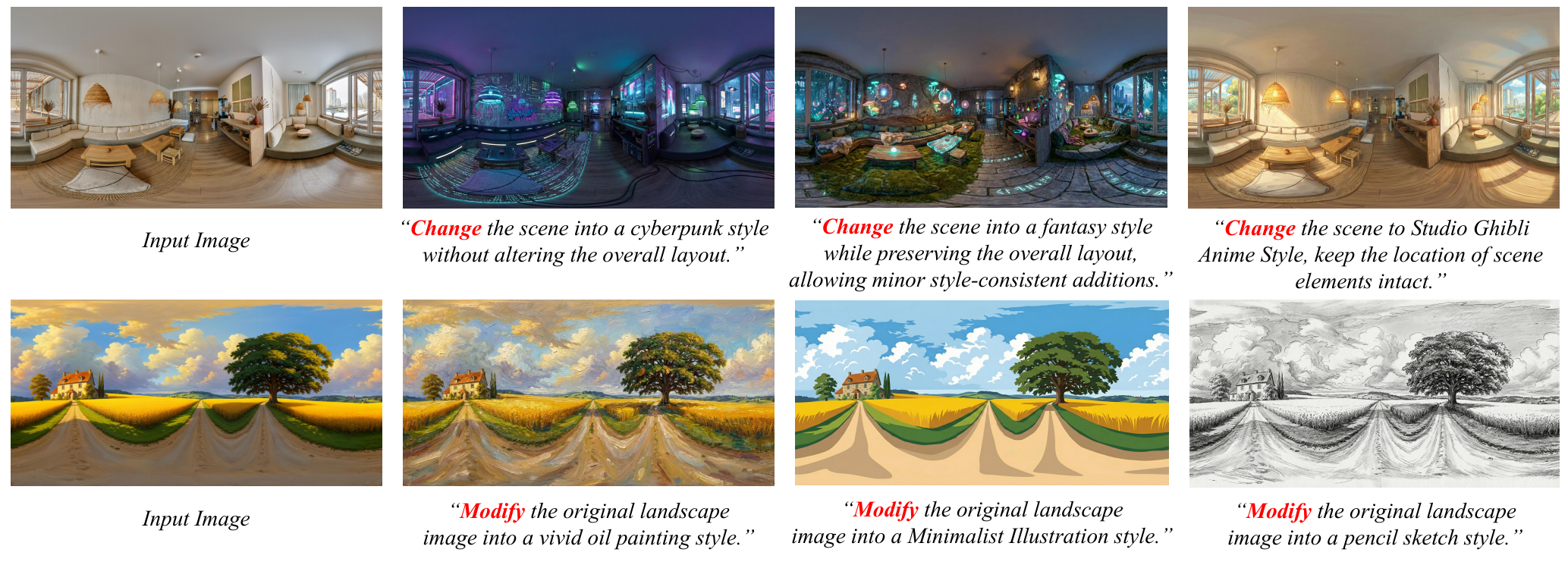}
    \caption{Global style editing results of \name.}
    \label{fig:global_style}
\end{figure*}

In addition to supporting local editing and simultaneous multi-object editing for panorama images, our method also facilitates global style editing, as shown in Fig.~\ref{fig:global_style}.

\subsection{More Visual Results}
\label{sec:more_results}
\begin{figure*}[t]
\centering
\includegraphics[width=1.0\linewidth]{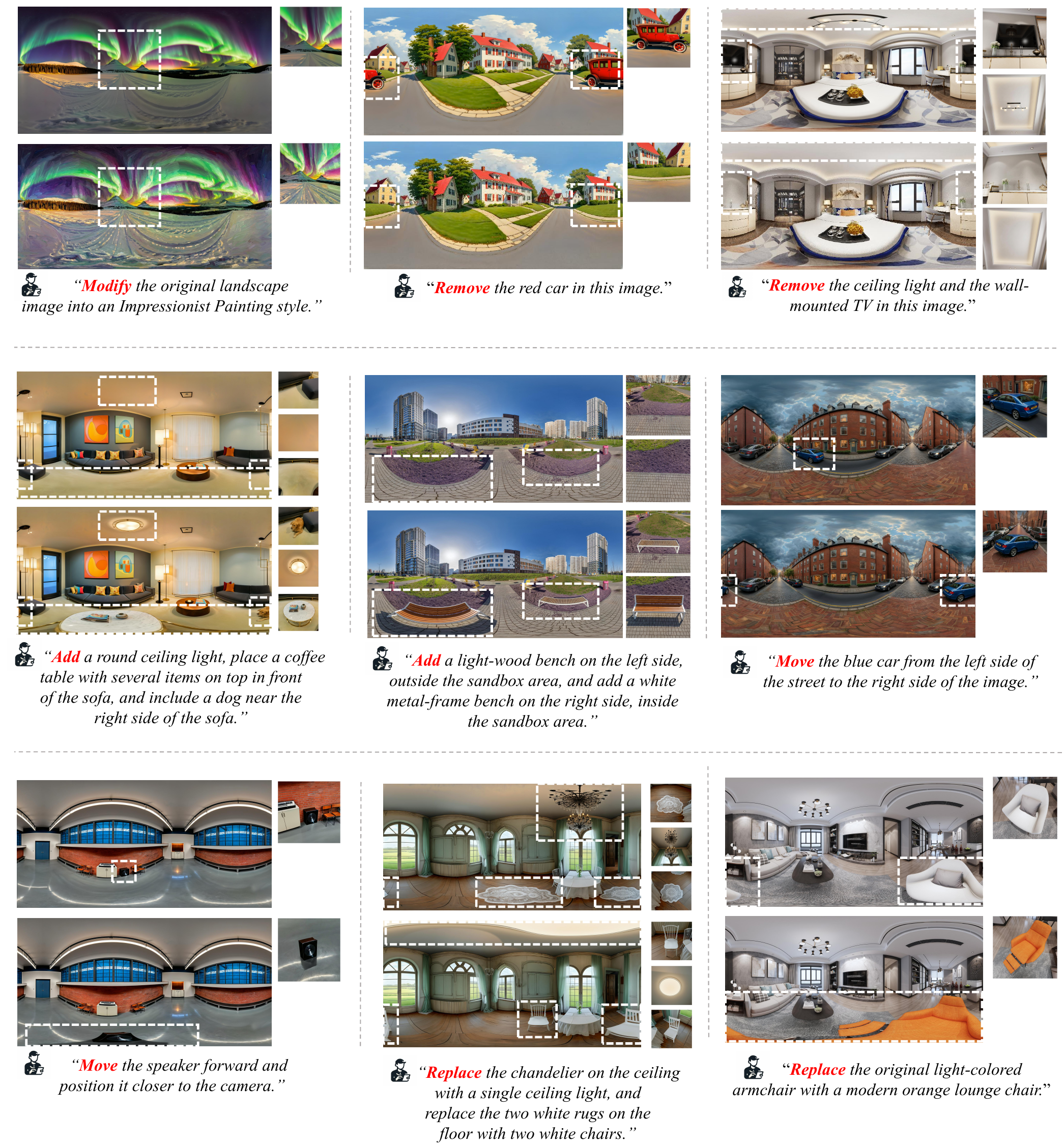}
    \caption{More editing results of our \name ~are shown below. For each case, the corresponding local perspective views are displayed in the right column, with the first row showing the results before editing and the second row showing the results after editing.}
    \label{fig:more_visual}
    \vspace{-3mm}
\end{figure*}

We provide more visual results in \figref{fig:more_visual}.

\section{Application}
\label{supp:application}

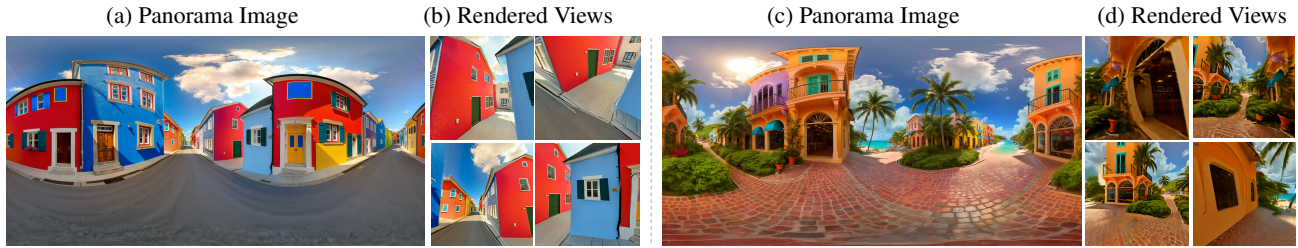
\begin{figure*}[t]
    \centering
\includegraphics[width=1.0\linewidth]{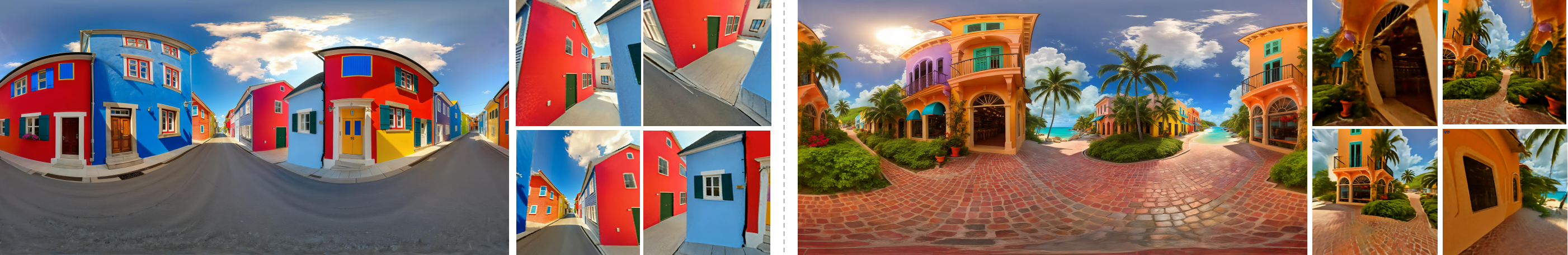}
    \put(-450,84){\small{(a) Panorama Image}}
    \put(-330,84){\small{(b) Rendered Views}}
    \put(-200,84){\small{(c) Panorama Image}}
    \put(-75,84){\small{(d) Rendered Views}}
    \caption{In this application, we transform a static panorama image into a 3D world environment. For each scenario, the input panorama image is displayed on the left, while the rendered views from the reconstructed 3D world appear in the four grids on the right.}
    \label{fig:app_1}
    \vspace{-3mm}
\end{figure*}

\begin{figure*}[t]
\centering
\includegraphics[width=1.0\linewidth]{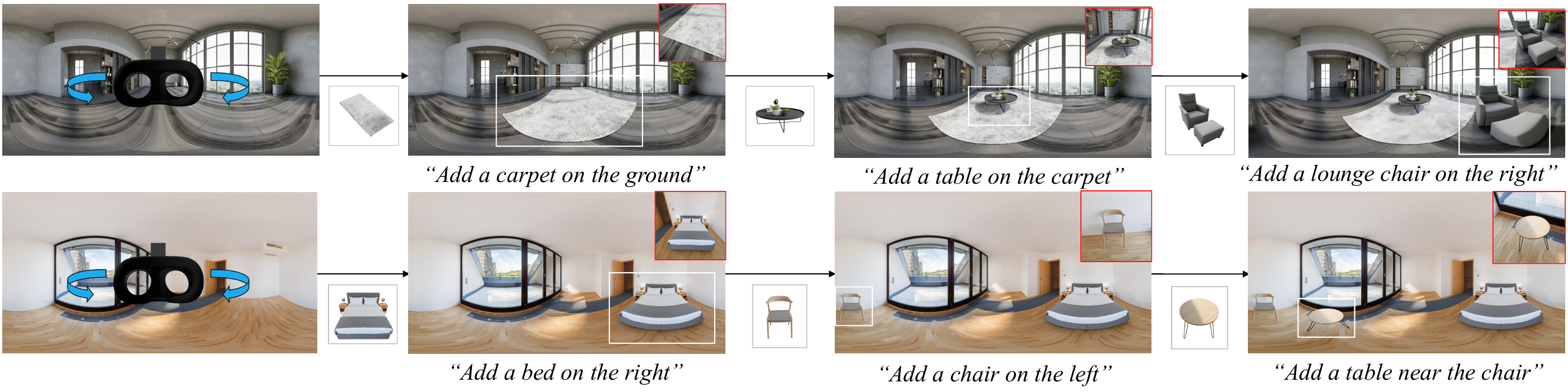}
\caption{In this application, users can experience a fully immersive 360-degree view of interior decoration through a VR headset.}
\label{fig:app_2}
\vspace{-3mm}
\end{figure*}

Our method supports diverse applications, and we show the visual results in \figref{fig:app_1} and \figref{fig:app_2}.

\noindent \textbf{\textit{3D world Generation}}: Users can begin by generating a panorama from either a text prompt or a local-view input image. A pre-trained depth estimation method~\cite{wang2025moge} is then applied to obtain the corresponding depth map of the panorama. Using this depth information, the 2D pixels are lifted into 3D points, and a sequence of camera poses is defined. The panorama is then rendered along these camera trajectories, and our method is employed to inpaint any missing regions in the rendered views. Finally, a panoramic Gaussian Splatting (GS) representation is optimized using the inpainted panorama frames.

\noindent \textbf{\textit{Indoor Design}}: Users fetch a desired piece of furniture from a catalog and specify a location in the room; our method then seamlessly integrates the object into the panoramic scene, adapting to the spherical geometry and lighting conditions for a photorealistic visualization.

\nocite{langley00}

\end{document}